\documentclass[11pt]{article}
\usepackage[utf8]{inputenc}
\usepackage[whole]{bxcjkjatype}

\usepackage{geometry}
\usepackage{fourier}
\usepackage{authblk}
\usepackage{bm}
\usepackage{tikz}
\usepackage{amsmath,amssymb,amsthm,bbm}
\usepackage{mathtools}

\usepackage{enumerate}
\usepackage{subcaption}
\usepackage{algorithm}
\usepackage{algorithmicx}
\usepackage{algpseudocode}
\usepackage{graphicx}[dvipdfmx]
\usepackage{mdframed}
\usepackage{natbib}
\usepackage{hyperref}
\usepackage{lipsum}
\hypersetup{
    colorlinks = true,
	citecolor=blue,
	linkcolor=blue
}

\usepackage{booktabs}

\theoremstyle{definition}
\newtheorem{theorem}{Theorem}
\newtheorem*{theorem*}{Theorem}
\newtheorem{proposition}{Proposition}
\newtheorem{definition}[theorem]{Definition}

\newtheorem{remark}[theorem]{Remark}
\newtheorem*{note*}{Note}

\usepackage{mdframed}

\newcommand{\ty}{\widetilde{y}}
\newcommand{\relu}{\text{ReLU}}
\newcommand{\cov}{\omega}
\newcommand{\COV}{\Omega}
\newcommand{\num}{\mathrm{num}}
\newcommand{\den}{\mathrm{den}}

\usepackage{titlesec}  
\usepackage{chngcntr}  
\usepackage{stmaryrd}
\usepackage{placeins}   

\newcommand{\ip}[2]{\left\llbracket \, #1 \, , \, #2 \, \right\rrbracket} 

\newcommand{\ideal}[1]{\left\langle #1 \right\rangle} 

\newcommand{\elim}[1]{\llparenthesis #1 \rrparenthesis}

\allowdisplaybreaks[4]

\title{Algebraic Approach to Ridge-Regularized \\ Mean Squared Error Minimization in Minimal ReLU Neural Network}
\author[1]{Ryoya Fukasaku\thanks{\href{mailto:fukasaku@math.kyushu-u.ac.jp}{fukasaku@math.kyushu-u.ac.jp}}}
\author[2]{Yutaro Kabata\thanks{\href{mailto:kabata@sci.kagoshima-u.ac.jp}{kabata@sci.kagoshima-u.ac.jp}}}
\author[3,4,5]{Akifumi Okuno\thanks{\href{mailto:okuno@ism.ac.jp}{okuno@ism.ac.jp} (corresponding author)}}
\affil[1]{Kyushu University}
\affil[2]{Kagoshima University}
\affil[3]{Institute of Statistical Mathematics}
\affil[4]{The Graduate University for Advanced Studies, SOKENDAI}
\affil[5]{RIKEN}
\date{\empty}

\begin{document}

\maketitle

\begin{abstract}
This paper investigates a perceptron, a simple neural network model, with ReLU activation and a ridge-regularized mean squared error (RR-MSE). Our approach leverages the fact that the RR-MSE for ReLU perceptron is piecewise polynomial, enabling a systematic analysis using tools from computational algebra. In particular, we develop a Divide-Enumerate-Merge strategy that exhaustively enumerates all local minima of the RR-MSE. 
By virtue of the algebraic formulation, our approach can identify not only the typical zero-dimensional minima (i.e., isolated points) obtained by numerical optimization, but also higher-dimensional minima (i.e., connected sets such as curves, surfaces, or hypersurfaces). 
Although computational algebraic methods are computationally very intensive for perceptrons of practical size, as a proof of concept, we apply the proposed approach in practice to minimal perceptrons with a few hidden units. 
\end{abstract}



\section{Introduction}

Neural networks~\citep{rosenblatt1958perceptron,jain1996artificial} are powerful predictive models that have driven major breakthroughs in computer vision~\citep{alexnet,dosovitskiy2021an}, language modeling~\citep{brown2020gpt,chang2024llm}, scientific applications~\citep{jumper2021highly,abramson2024accurate}, and engineering~\citep{reed2022generalist,zitkovich2023rt2} over the past decade, largely due to advances in computing power and the availability of large-scale data~\citep{LeCun2015,Goodfellow-et-al-2016}. Neural networks come in various forms, including convolutional neural networks~\citep{fukushima1980neocognitron,lecun1989backpropagation,alexnet}, graph neural networks~\citep{kipf2017semisupervised}, and Transformers~\citep{vaswani2017attention}, among others. Despite their architectural diversity, most neural networks share a common foundational structure, with the multilayer perceptron~\citep{rosenblatt1958perceptron,bishop1995neural} serving as a simple yet fundamental building block.

The impressive performance of neural networks has led to active research on both their theoretical and empirical foundations. The universal approximation theorem~\citep{Cybenko1989,Hornik1989,Funahashi1989} shows that even shallow networks can approximate any continuous function, and it has been extended to deeper architectures~\citep{elbrachter2021deep,shen2022optimal}. Several studies have examined generalization through norm-based complexity measures~\citep{Bartlett2017Spectrally,Neyshabur2018Towards} and other studies investigated the geometry of loss landscapes~\citep{Keskar2017LargeBatch,dinh2017sharp}. Optimization theory suggests that saddle points, rather than local minima, pose the primary challenge in training~\citep{dauphin2014identifying}, and under certain conditions, all local minima may be globally optimal~\citep{kawaguchi2016deep,kawaguchi2019every}. On the empirical side, for instance, visualizations of the loss surface reveal broad, flat valleys associated with better generalization~\citep{LiLossLandscape2018}, and spectral analyses have shown that Hessians typically exhibit heavy-tailed spectra, with only a few directions contributing significant curvature~\citep{Ghorbani2019Hessian}. These mainstream approaches, which adopt a top-down perspective focusing on global properties such as generalization and loss geometry, have significantly advanced our understanding of neural network behavior.

As a contrasting approach to the top-down perspective, another line of research seeks to understand neural networks by analyzing small-scale models from a bottom-up viewpoint. This direction often centers on perceptrons with ReLU activations and explores their algebraic structure through mathematical analysis. 
In this line, as an early contribution to the algebraic study of neural networks, \citet{Montufar2014} and \citet{charisopoulos2019tropical} analyzed the number of linear regions in deep networks, relating it to expressive power of the deep perceptron. The piecewise-linear behavior of such perceptrons are further investigated by tropical geometry~\citep{maclagan2015introduction}, rooted in max-plus algebra. 
\citet{zhang2018tropical} showed a precise equivalence between ReLU perceptrons and tropical rational maps, offering a geometric view via tropical hypersurfaces. \citet{alfarra2022decision} examined decision boundaries through tropical algebra, focusing on the effects of pruning and initialization, and linking to the lottery ticket hypothesis. \citet{brandenburg2024the} studied these boundaries through tropical rational functions and polyhedral fan subdivisions. Beyond standard perceptrons, tropical geometry has been applied to graph neural networks~\citep{pham2024tropicalGNN}, and surveys such as \citet{maragos2021tropical} outline how tropical and max-plus methods unify perspectives across neural networks.

These studies offer valuable insights into the decision boundaries of ReLU-activated perceptrons. However, they yield limited implications for the loss landscape or the global geometry of the parameter space, prompting growing interest in algebraic approaches to loss surface analysis. Among them, \citet{mehta2022} applies computational algebraic approach to deep linear perceptrons, which does not apply any activation function; \citet{mehta2022} lists all the solutions in the multilayer linear model. The study aligns closely with our own in its algebraic approach to enumerate all local minima, although it remains limited to non-activated perceptron.

Although omitting activations greatly simplifies the perceptron and facilitates mathematical analysis, this simplification significantly departs from the behavior of ReLU perceptrons. In particular, it fails to capture essential features such as the partitioning of the parameter space into polyhedral regions and the resulting boundary phenomena. To bridge this gap, we develop a computational algebraic framework that exhaustively enumerates local minima in ReLU perceptrons. Our method exploits the fact that ridge-regularized mean squared error for ReLU perceptrons is piecewise polynomial, allowing systematic treatment via computational algebra. 
By virtue of the algebraic formulation, our approach can identify not only the typical zero-dimensional minima (i.e., isolated points) obtained by numerical optimization, but also higher-dimensional minima (i.e., connected sets such as curves, surfaces, or hypersurfaces). It also enables the exhaustive enumeration of all local minima. 
While algebraic methods are typically too intensive for large networks, we apply the proposed algorithm to minimal ReLU perceptrons with few hidden units as a proof of concept.

\bigskip

The remainder of this paper is organized as follows.
Section~\ref{subsec:related_works} reviews remaining related works.
Section~\ref{subsec:problem_settings} presents the problem setting.
Section~\ref{sec:preparation} introduces preparatory steps before leveraging computational algebra.
Section~\ref{sec:proposed_algorithm} describes the proposed algebraic algorithm.
Section~\ref{sec:numerical} provides numerical demonstrations.
Finally, Section~\ref{sec:conclusion} concludes the paper.

\subsection{Related Works}
\label{subsec:related_works}

For one-hidden-layer perceptrons with ReLU activation, \citet{arora2018understanding} proposed a brute-force approach that partitions the parameter space into polyhedral regions and exhaustively searches for the global optimum. However, enumerating all such regions is computationally prohibitive; in fact, \citet{boob2022complexity} formally established that training ReLU perceptrons is NP-hard. Although the original loss function is non-convex, \citet{pilanci2020neural} demonstrated that the training problem can be exactly reformulated as an equivalent convex optimization problem at the expense of an increased number of parameters, which scales with the number of polyhedral regions. By introducing gated ReLU activation~\citep{fiat2019decoupling}, \citet{mishkin2022fast} efficiently approximated the convex reformulation, and further demonstrated that the approximation can be regarded as a constrained group Lasso~\citep{mishkin2023optimal}. Another convex relaxation through \citet{pilanci2020neural} is explored in \citet{kim2024convex}; see the detailed survey therein. While these methods aim to convexify the original non-convex loss and recover a global optimum, our approach takes a different route by algebraically enumerating all local minima, rather than focusing on a single global solution.

Singular learning theory, developed by \citet{watanabe2009}, analyzes the local geometric structure of parameter spaces in singular models and its influence on Bayesian generalization error~\citep{watanabe2010,drton2017bayesian}. This theory characterizes generalization behavior using several algebraic invariants; these quantities are analytically computed for various models, including reduced-rank regression~\citep{aoyagi2005stochastic}, restcirted Boltzmann machines~\citep{aoyagi2010boltzmann}, three-layer neural networks~\citep{aoyagi2019vandermonde}, deep linear networks~\citep{aoyagi2024consideration}, and generalized semiregular models~\citep{kurumadani2025learning}, and so forth. These works primarily focus on understanding and quantifying generalization the local behavior of singular models.

Algebraic composition of neurons in neural networks has been explored from various perspectives. \citet{ritter2003lattice} proposed a neural network framework based on lattice algebra, where numerical operations are replaced by order-theoretic and logical ones to support logic-driven, non-numeric inference. \citet{peng2017multilayerperceptronalgebra} introduced MLP Algebra, defining algebraic operations (sum, product, complement) for systematically constructing complex networks from simpler modules. \citet{cruz2018neural} developed a neural system that composes classifiers using Boolean logic (AND, OR, NOT), enabling zero-shot recognition of complex visual concepts. More recently, \citet{parada-mayorga2021algebraic} presented a unifying framework based on algebraic signal processing, demonstrating improved stability in neural architectures through structural properties grounded in commutative algebra.

Algebraic structuring of internal representations in neural networks replaces real-valued computations with richer algebraic frameworks. \citet{Buchholz2001} and \citet{Buchholz2008clifford} introduced Clifford algebra perceptrons, extending standard perceptrons to operate over Clifford algebras. \citet{li2019extended} further generalized this framework to reduced geometric algebras, enabling more efficient and expressive processing of multi-dimensional signals. Another developments include complex-valued~\citep{hirose1992proposal}, quaternion-valued~\citep{arena1997quaternion}, and $C^*$-algebra based~\citep{Hashimoto2021} neural networks.

Computational algebra has recently begun to be applied to the analysis of statistical methods. For instance, \citet{fukasaku2024algebraic} surveys the algebraic structure of factor analysis, and \citet{fukasaku2025algebraic} further investigates orthomax rotations used in factor analysis.



\subsection{Problem Settings}
\label{subsec:problem_settings}

In this section, we present the formal problem setting considered in this study. 

Let \( n, d \in \mathbb{N} \) represent the sample size and the dimensionality of the covariates, respectively.  
The data consist of fixed observations \( \{(x_i, y_i)\}_{i=1}^{n} \subset \mathbb{R}^{d} \times \mathbb{R}\).  
We aim to predict the observed outcome \( y_i \in \mathbb{R} \) from the covariate \( x_i \in \mathbb{R}^d \) using a neural network. In particular, we focus on a special case of neural networks known as the one-hidden-layer perceptron~\citep{rosenblatt1958perceptron,bishop1995neural}, defined by the function:
\begin{align}
    \mathbb{R}^d \ni 
    x \mapsto 
    \ip{a}{\relu(Bx + c)} + m \in \mathbb{R},
\label{eq:single-layer-perceptron}
\end{align}
where \( a \in \mathbb{R}^L, B=(b_1^{\top},b_2^{\top},\ldots,b_L^{\top})^{\top} \in \mathbb{R}^{L \times d}, c=(c_1,c_2,\ldots,c_L) \in \mathbb{R}^L, m \in \mathbb{R} \) are parameters, \( L \) denotes the number of hidden units, and the rectified linear unit \( \relu(z) \) applies the operation \( z \mapsto \max\{0, z\} \) element-wise. $\ip{a}{b}$ denotes the inner product between two vectors $a$ and $b$.

Within these parameters, the bias term \( m \in \mathbb{R} \) can be eliminated by centering the outcomes \( \{y_i\} \).  
Equipped with the centered outcomes \( \widetilde{y}_i = y_i - n^{-1} \sum_{i=1}^{n} y_i \), we consider minimizing the ridge-regularized mean squared error (RR-MSE):
\begin{align}
    \widetilde{\ell}_{\lambda}(a,B,c)
    =
    \sum_{i=1}^{n} \left\{ \widetilde{y}_i - \ip{a}{\relu(Bx_i + c)} \right\}_2^2
    + \lambda \left( \|a\|^2 + \|B\|^2 + \|c\|^2 \right).
    \label{eq:R2MSE}
\end{align}
Here, \( \lambda > 0 \) serves as a regularization constant; it is specified a priori and not estimated from data in this study. 
Our objective is to identify the complete set of local minima of the RR-MSE using computational algebra, i.e., 
\begin{align}
\Theta^* = \{\theta = (a,B,c) \text{ is a local minimum of }\widetilde{\ell}_{\lambda}(a,B,c) \}
\subset \Theta = \mathbb{R}^{L} \times \mathbb{R}^{L \times d} \times \mathbb{R}^L,
\label{eq:goal}
\end{align}
where the local minimum is rigorously defined as follows. 

\begin{definition}
A point $\theta^*$ is said to be a local minimum of a function $\ell(\theta):\Theta \to \mathbb{R}$ if there exists $\varepsilon > 0$ such that $\ell(\theta^*) \le \ell(\theta)$ for all $\theta$ satisfying $\|\theta - \theta^*\|_2 \le \varepsilon$.
\end{definition}

Note that the local minimum defined above is not necessarily an isolated point; for example, every point along a continuous set of local minima, such as a valley, is also included in the definition of a local minimum. 
\( L \) is typically chosen to be a large number (e.g., \( L \sim 10^4 \)). However, in computational algebraic approaches, the computational cost increases dramatically with the number of variables. From a practical standpoint, we select small \( L \), which allows us to analyze essential properties such as the symmetry of hidden units under permutation, while keeping the number of units manageable.

\begin{figure}[!ht]
\centering
\includegraphics[width=0.95\textwidth]{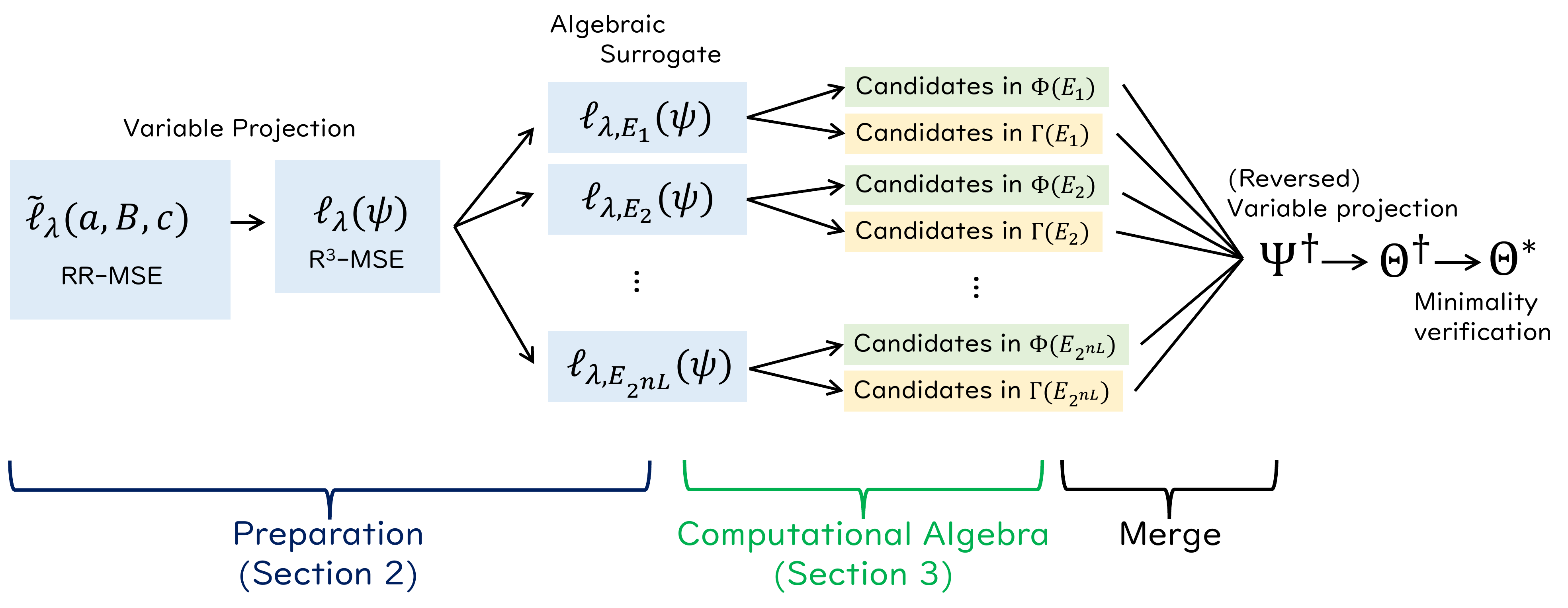}
\caption{Our method is based on the Divide-Enumerate-Merge strategy described in Section~\ref{subsec:DEM}. First, we reduce the number of parameters by transforming the original RR-MSE into the reduced RR-MSE (R$^3$-MSE). Second, the R$^3$-MSE is decomposed into algebraically tractable surrogate functions, each of which coincides with the original R$^3$-MSE on a specific partition $\Psi(E)$. Third, we exhaustively enumerate all local minimum candidates in each partition in parallel via computational algebra. Fourth, we merge the resulting candidate minima across partitions. Finally, we filter these candidates by verifying local minimality to obtain the exact set $\Theta^*$, which coincides with the complete set of local minima of the RR-MSE.}
\label{fig:process_overview}
\end{figure}

At the end of this section, we provide an overview of our approach in Figure~\ref{fig:process_overview}. Also, we provide Table~\ref{tab:symbols} as a helpful reference, summarizing the key symbols and their meanings for the reader's convenience.

\begin{table}[!ht]
\centering
  \caption{Key notations and their descriptions}
  \label{tab:symbols} 
\begin{tabular}{ll}
  \toprule
  Notation & Description \\
  \midrule 
  $L$ & Number of hidden units in the perceptron defined in \eqref{eq:single-layer-perceptron}. \\
  $n,d$ & Sample size and covariate dimension, respectively. \\
  $\widetilde{y}_i$ & Centered outcome, i.e., $\widetilde{y}_i=y_i-n^{-1}\sum_{i=1}^{n}y_i$. \\
  $\ip{a}{b}$ & Inner product between vectors $a$ and $b$. \\
  $\lambda>0$ & Ridge regularization constant. \\
  $\psi=(B,c)$ & Essential parameters (redundant parameter $a$ will be eliminated in Section~\ref{subsec:variable_projection}). \\
  $w=L(d+1)$ & Number of entries in the essential parameter $\psi$, i.e., parameter dimension. \\
  $\Psi = \mathbb{R}^{L \times d} \times \mathbb{R}^L$ & Entire parameter space. \\
  $\xi_{i\ell}(\psi)$ & Linear function $\ip{b_{\ell}}{x_i}+c_{\ell}$ defined in \eqref{eq:E_and_xi}. \\
  $E=(e_{i\ell}) \in \{\pm 1\}^{L \times d}$ & Indicator matrix. \\
  $\Psi(E)$ & Partition of the parameter space $\Psi$ defined in \eqref{eq:partition}. \\
  $\Phi(E)$ & Interior of the partition $\Psi(E)$. \\
  $\Gamma(E)$ & Boundary of the partition $\Psi(E)$. \\
  $\Gamma_{i\ell}(E)$ & Boundary component defined in \eqref{eq:boundary_component}, and it satisfies $\Gamma(E)=\bigcup_{i\ell}\Gamma_{i\ell}(E)$.  \\
  $\omega_{i}(\psi)$ & Concatenation of ReLU outputs of the linear function $\xi_{i\ell}(\psi)$, for $\ell=1,2,\ldots,L$. \\
  $\omega_{iE}(\psi)$ & Surrogate of $\omega_{i}(\psi)$ over the partition $\Psi(E)$, defined in \eqref{eq:ReLU_surrogate}. \\  
  $\widetilde{\ell}_{\lambda}(a,B,c)$ & Ridge regularized-MSE (RR-MSE) defined in \eqref{eq:R2MSE}. \\
  $\ell_{\lambda}(\psi)$ & Reduced RR-MSE (R$^3$-MSE) defined in \eqref{eq:R3MSE}. \\
  $\ell_{\lambda,E}(\psi)$ & Algebraic surrogate defined in \eqref{eq:algebraic_surrogate}. \\
  $\mathbb{R}[\psi]$ & Polynomial ring with real coefficients. \\
  $\mathcal{I}=\ideal{f_1,f_2,\ldots,f_r}$ & Ideal generated by polynomials $f_1,f_2,\ldots,f_r \in \mathbb{R}[\psi]$. \\
  $\mathbb{V}(f_1,f_2,\ldots,f_r)$ & Algebraic variety defined in \eqref{eq:algebraic_variety}, for polynomials $f_1,f_2,\ldots,f_r \in \mathbb{R}[\psi]$. \\
  $\mathrm{LM}(f)$ & Leading monomial in the polynomial $f \in \mathbb{R}[\psi]$. \\ 
  $\mathcal{I}:\mathcal{K}^{\infty}$ & The saturation ideal of the ideal $\mathcal{I}$ with respect to the ideal $\mathcal{K}$. \\
  $\elim{\mathcal{I}}_i$ & $i$th elimination ideal of the ideal $\mathcal{I}$. \\
  $\num(r(\psi)),\den(r(\psi))$ & Numerator and denominator of the rational function $r(\psi)$, respectively. \\
  $\Psi^{\dagger} \subset \Psi$ & A set of local minima candidates of R$^3$-MSE. \\
  $\Theta^{\dagger}$ & A set of local minima candidates of RR-MSE. \\
  $\Theta^*$ & Complete set of exact local minima of RR-MSE. \\
  \bottomrule
\end{tabular}
\end{table}

\section{Preparation for Algebraic Computation}
\label{sec:preparation}

To facilitate the use of computational algebra, both from the standpoint of computational tractability and algebraic compatibility, it is essential to reduce the number of parameters and reformulate the RR-MSE into a structure amenable to algebraic manipulation.  To this end, this section introduces several preparatory techniques, called variable projection (in Section~\ref{subsec:variable_projection}) and algebraic surrogate (in Section~\ref{subsec:algebraic_surrogate}).

\subsection{Variable Projection}
\label{subsec:variable_projection}

In computational algebra, the computational complexity increases significantly with the number of parameters.  
In our problem setting, however, only \( (B, c) \) are essential, as the parameter \( a \) can be analytically removed via the variable projection technique (see, e.g., \citet{golub1973differentiation} and \citet{golub2003separable}).  
A conceptually similar approach can be found in the profile likelihood framework in statistics (see, e.g., \citet{murphy2000profile}). 

The explicit variable projection procedure is outlined below.
Let the essential parameters be
\[
\psi = (B, c) \in \Psi = \mathbb{R}^{L \times d} \times \mathbb{R}^L.
\]
Fixing \( \psi \), the optimal parameter \( a \) is obtained by solving a linear regression problem with response variables \( \{\widetilde{y}_i\} \) and covariates \( \cov_i = \cov_i(\psi) := \relu(Bx_i + c) \).  
Let \( \COV = \COV(\psi) = (\cov_1^{\top}, \cov_2^{\top}, \ldots, \cov_n^{\top})^{\top} \in \mathbb{R}^{n \times L} \).  
Then, the RR-MSE is uniquely minimized by
\[
\hat{a}(\psi) = (\COV^{\top} \COV + \lambda I_L)^{-1} \COV^{\top} \widetilde{y}.
\]
Here, we adopt the variable projection technique, originally introduced in the unpenalized setting by \citet{golub1973differentiation} and \citet{golub2003separable}, and later extended to penalized problems by \citet{chen2019regularized} and \citet{espanol2023variable}. Variable projection substitutes $\hat{a}(\psi)$ back into the RR-MSE, and it yields its (locally) minimal value, which coincides with the reduced RR-MSE (R$^3$-MSE):
\begin{align}
    &\ell_{\lambda}(\psi) 
    =
    \lambda \|\psi\|^2
    -
    \ip{\widetilde{y}}{
        \COV(\COV^{\top} \COV + \lambda I_L)^{-1} \COV^{\top} \widetilde{y}
    }
    \label{eq:R3MSE} 
\end{align}
up to a constant term \( \|\widetilde{y}\|_2^2 \). That is, $\ell_{\lambda}(\psi) = \min_{a \in \mathbb{R}^L} \widetilde{\ell}_{\lambda}(a,\psi) - \|\widetilde{y}\|_2^2$. \( I_L \) denotes the \( L \times L \) identity matrix.  
The full derivation is provided in Appendix~\ref{app:derivation_of_eq:R3MSE}.

With any fixed $\psi \in \Psi$, the RR-MSE admits a unique minimizer \( \hat{a}(\psi) \), since it is strongly concave with respect to \( a \). 
As a result, the set of local minima of the RR-MSE coincides with the set of local solutions obtained by minimizing the R$^3$-MSE with respect to \( \psi \) (and the corresponding \( \hat{a}(\psi) \)).  
Accordingly, our objective is reduced to the task of locally minimizing the R$^3$-MSE with respect to $\psi$, and to that end, we seek to enumerate the full set of its local minima using tools from computational algebra. More specifically, we develop an algorithm that identifies a ``rough'' set of local minima candidates $\Psi^{\dagger} \subset \Psi$ such that
\[
    \left\{ \psi=(B,c) \text{ is a local minimum of } \ell_{\lambda}(\psi) \right\} \subset \Psi^{\dagger} \subset \Psi,
\]
and applying the reversed variable projection to each element of $\Psi^{\dagger}$ yields the set $\Theta^{\dagger}$ satisfying
\begin{align}
    \Theta^* 
    \subset 
    \Theta^{\dagger} = \left\{ (\hat{a}(\psi), \psi) \mid \psi \in \Psi^{\dagger} \right\}.
    \label{eq:reversed_varible_projection}
\end{align}
Subsequently, we verify the local minimality of each element in $\Theta^{\dagger}$ and discard non-minimizing stationary points to achieve our overarching goal, i.e., the exact complete set of local minima $\Theta^*$. Accordingly, the remainder of this paper is devoted to the construction of $\Psi^{\dagger}$.

\subsection{Algebraic Surrogate Functions}
\label{subsec:algebraic_surrogate}

\begin{figure}[!t]
\centering
    \includegraphics[width=0.9\textwidth]{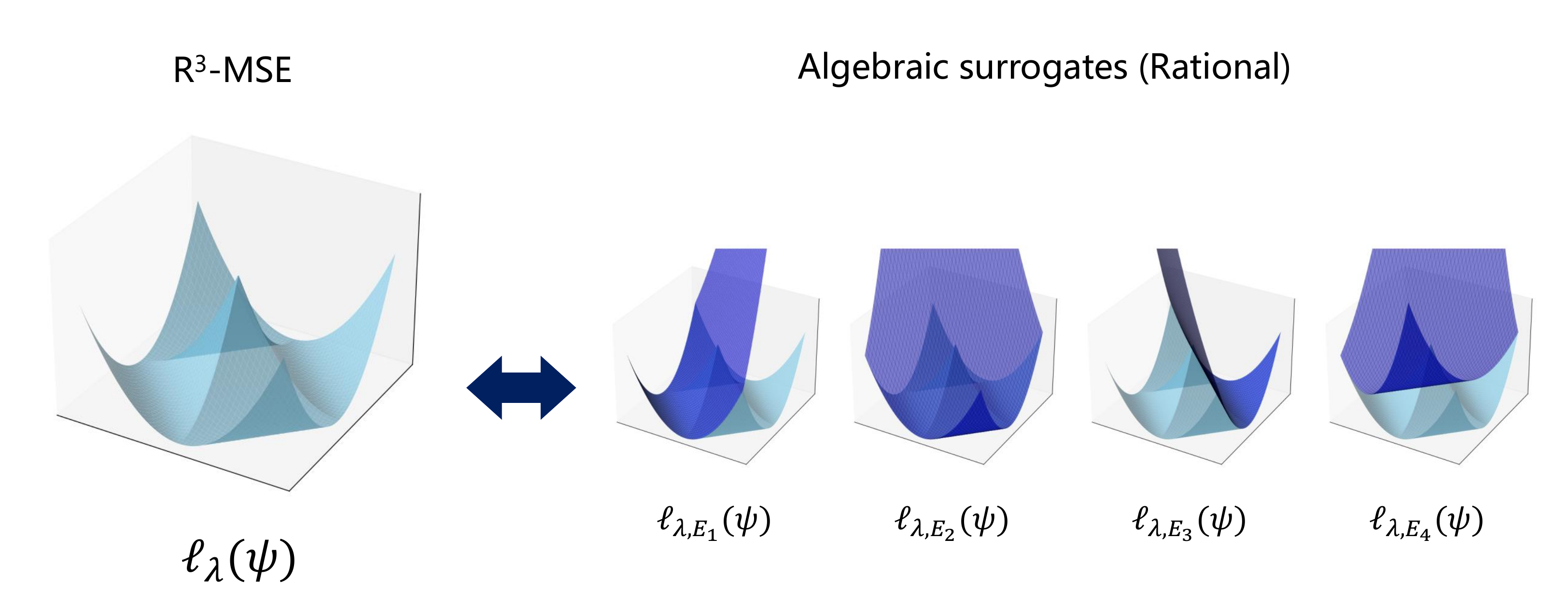}
    \caption{Concept of the algebraic surrogate function \( \ell_{\lambda,E}(\psi) \): the surrogate coincides with the original R$^3$-MSE \( \ell_{\lambda}(\psi) \) within the partition \( \Psi(E) \subset \Psi \), while maintaining algebraic tractability.}
    \label{fig:illustration_surrogates}
\end{figure}

While the RR-MSE is a piecewise polynomial function, its polynomial form depends on the parameter region due to the presence of ReLU activations. As a result, applying computational algebra to the RR-MSE (and the R$^3$-MSE) in the entire region globally is inherently challenging. To overcome this difficulty, we introduce the concept of algebraic surrogate functions, which serve as a globally tractable alternative to the R$^3$-MSE. An illustrative example of this concept is shown in Figure~\ref{fig:illustration_surrogates}.

Define
\begin{align}
    E = (e_{i \ell}) \in \{\pm 1\}^{n \times L},
    \quad 
    \text{ and }
    \quad 
    \xi_{i\ell}(\psi) = \ip{b_{\ell}}{x_i} + c_{\ell}
    \label{eq:E_and_xi}
\end{align}
as an indicator matrix and linear function, respectively, for $i=1,2,\ldots,n$ and $\ell=1,2,\ldots,L$. This indicator matrix and the linear function divides the parameter set $\Psi$ into partitions defined by
\begin{align}
    \Psi(E)
    =
    \left\{
    \psi \in \Psi 
    \mid 
    \xi_{i\ell}(\psi) e_{i\ell} \ge 0, \, \forall i, \ell
    \right\}.
    \label{eq:partition}
\end{align}
The indicator matrix $E$ is used to represent the activation patterns of the ReLU hidden units. For any parameter $\psi \in \Psi(E)$, the linear function $\xi_{i\ell}(\psi)$ is active when $e_{i\ell} = 1$ and inactive when $e_{i\ell} = -1$. A similar concept that exploits all possible activation patterns can be found in the existing literature. For instance, \citet{pilanci2020neural}, \citet{mishkin2022fast}, and \citet{mishkin2023optimal} consider activation patterns to reformulate the training of ReLU neural networks (with sufficiently many hidden units) as a convex problem.

We can then easily prove the following proposition, which states that \( \{\Psi(E)\} \) form (partially-overlapped) convex regions that partition the parameter set \( \Psi \).

\begin{proposition}\label{pro:divprm}
Let \( E \in \{\pm 1\}^{n \times L} \). The following properties hold: 
(1) \( \Psi = \bigcup_{E \in \{\pm 1\}^{n \times L}} \Psi(E) \). 
(2) For any \( \psi, \tilde{\psi} \in \Psi(E) \) and \( \alpha \in [0, 1] \). 
(3) \( \alpha \psi + (1 - \alpha) \tilde{\psi} \in \Psi(E) \). 
(4) For any \( \psi \in \Psi(E) \) and \( \beta \ge 0 \),  
    \( \beta \psi \in \Psi(E) \).
(5) \( \Psi(-E) = -\Psi(E) \).
\end{proposition}

We emphasize that the convex partition \( \{\Psi(E)\} \) exhibits partial overlap. For instance, for any \( E \in \{\pm 1\}^{n \times L} \), it holds for \( E' = -E \) that $\Psi(E) \cap \Psi(E') = \{\psi \in \Psi \mid \exists i, \ell, \, \text{s.t.} \, \xi_{i\ell}(\psi) e_{i\ell} = 0\} \neq \emptyset$.

\bigskip
The R$^3$-MSE is a continuous rational function over the parameter space \( \Psi \). It is differentiable on both sides in the interior of each partition \( \Psi(E_i) \), but not across the boundaries between partitions. Therefore, to identify all local minima, it is necessary to further introduce a new representation of the $R^3$-MSE. In the following, we construct such a representation; herein, we first provide a redundant but algebraically tractable expression for the perceptron with ReLU activation~\eqref{eq:single-layer-perceptron}.

Let \( E = (e_{i\ell}) \in \{\pm 1\}^{n \times L} \) be a fixed indicator matrix. Define the ReLU output \( \omega_i = \omega_i(\psi) = \relu(Bx_i + c) \) and its surrogate
\begin{align}
  \cov_{iE} = \cov_{iE}(\psi) := 
  \left(
  \frac{e_{i\ell} + 1}{2} \, \xi_{i\ell}(\psi)
  \right) \in \mathbb{R}^L.
    \label{eq:ReLU_surrogate}
\end{align}
Then, for any parameter \( \psi \in \Psi(E) \), this surrogate expression recovers the original ReLU output:
\[
  \cov_{iE} = \cov_i,
\]
since $\frac{e_{i\ell}+1}{2}\xi_{i\ell}(\psi)=\mathbbm{1}(e_{i\ell}=+1)\xi_{i\ell}(\psi)=\mathbbm{1}(\xi_{i\ell}(\psi)\ge 0)\xi_{i\ell}(\psi)=\relu(\xi_{i\ell}(\psi))$ holds for all \( i, \ell \). 
This formulation eliminates the need to explicitly invoke the ReLU function; over the region \( \Psi(E) \), the output of the perceptron~\eqref{eq:single-layer-perceptron} can be represented by a simple polynomial.  
As a result, we are able to effectively leverage computational algebra for further analysis.

With \(\COV_E = \COV_E(\psi)= (\cov_{1E}^{\top}, \cov_{2E}^{\top}, \ldots, \cov_{nE}^{\top})^{\top} \in \mathbb{R}^{n \times L}\), we also define a rational function
\begin{align}
    \ell_{\lambda,E}(\psi) 
    =
    \lambda \|\psi\|^2
    -
    \ip{\widetilde{y}}{\COV_E(\COV_E^{\top} \COV_E + \lambda I_L)^{-1} \COV_E^{\top} \widetilde{y}},
    \label{eq:algebraic_surrogate}
\end{align}
where we call it as {\em  algebraic surrogate function}. 
As $\Omega_E(\psi)=\Omega(\psi)$ holds over the partition $\Psi(E)$, this algebraic surrogate also coincides with the R$^3$-MSE, i.e., 
\[
    \ell_{\lambda,E}(\psi) = \ell_{\lambda}(\psi), 
    \, 
    \text{ for any }
    \, 
    \psi \in \Psi(E).
\]
An illustration of the surrogate is provided in Figure~\ref{fig:illustration_surrogates} and further exemplified in Supplement~\ref{supp:illustrative_example}. 
The key advantage of this formulation lies in its rational structure, which allows us to fully exploit the power of computational algebra. This enables the complete enumeration of all local minima in the surrogate over the entire parameter space \( \Psi = \mathbb{R}^{L \times d} \times \mathbb{R}^L \).

\begin{figure}[!t]
\centering
\begin{minipage}{0.45\textwidth}
    \includegraphics[width=\textwidth]{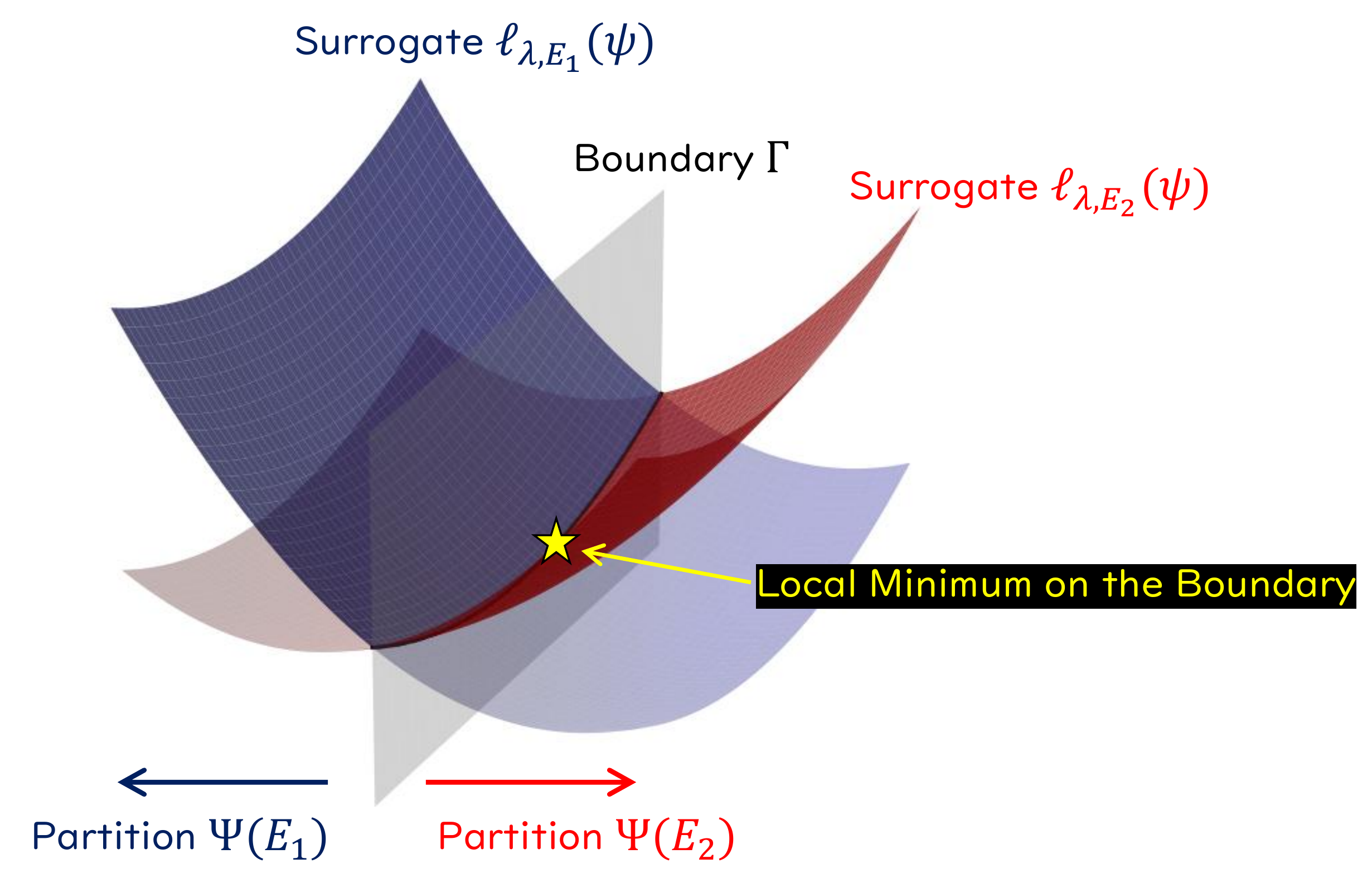}
    \subcaption{Local minimum in the unconstrained domain $\Psi$.}
    \label{subfig:minimum_in_unconstrained_domain}
\end{minipage}
\begin{minipage}{0.45\textwidth}
    \includegraphics[width=\textwidth]{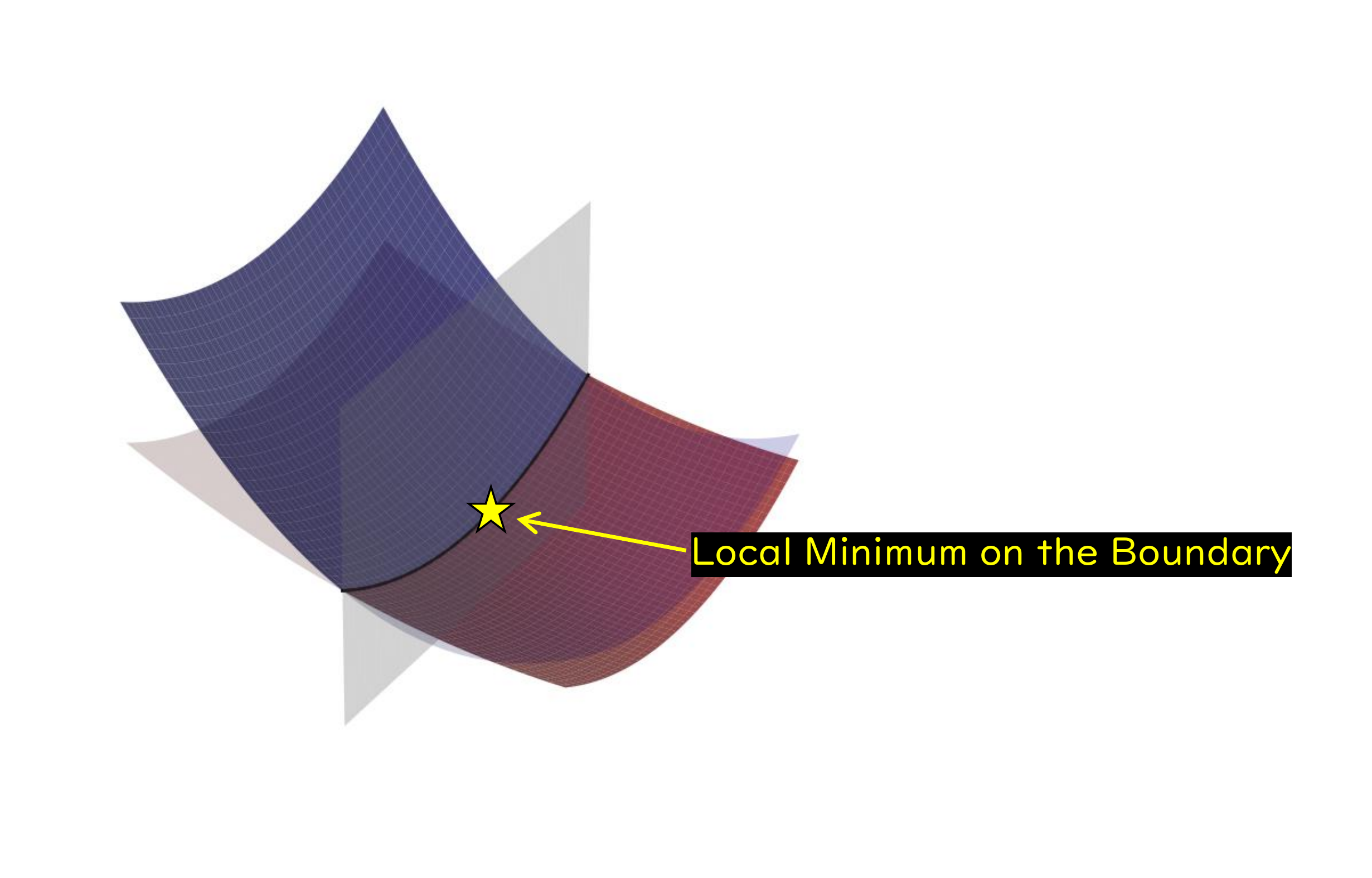}
    \subcaption{Counterexample.}
    \label{subfig:counterexample}
\end{minipage}
    \caption{The boundary local minimum of $\ell_{\lambda}(\psi)$ is plotted along with the algebraic surrogate functions $\ell_{\lambda,E_j}(\psi)$ for $j = 1, 2$. As stated in Theorem~\ref{thm:main} and illustrated in (\subref{subfig:minimum_in_unconstrained_domain}), any local minimum in the unconstrained domain $\Psi$, which also belongs to the boundary, must also be a local minimum restricted to the boundary $\Gamma$. However, the converse does not hold: a local minimum on the boundary is not necessarily a local minimum in the full domain, as illustrated in (\subref{subfig:counterexample}).}
    \label{fig:illustration_boundary_minimum}
\end{figure}

Roughly speaking, the local minima of the algebraic surrogate \( \ell_{\lambda,E}(\psi) \) coincide with those of the R$^3$-MSE \( \ell_{\lambda}(\psi) \) within the interior of the partition \( \Psi(E) \). However, optimization over partition boundaries entails additional complexity. If a point on the boundary is a local minimum of the R$^3$-MSE \( \ell_{\lambda}(\psi) \) in a neighborhood (not restricted to the boundary), then it also remains a local minimum of the algebraic surrogate \( \ell_{\lambda,E}(\psi) \) when considering neighborhoods constrained to the boundary. The converse, however, does not necessarily hold. These results are illustrated in Figure~\ref{fig:illustration_boundary_minimum}, and rigorously established in Theorem~\ref{thm:main}.

To state Theorem~\ref{thm:main} precisely, we define the following boundary component:
\begin{align}
    \Gamma_{i\ell}(E) = \{\psi \in \Psi(E) \mid \xi_{i\ell}(\psi)=0 \},
    \label{eq:boundary_component}
\end{align}
for each \( i = 1, \ldots, n \) and \( \ell = 1, \ldots, L \). The overall boundary of the partition $\Psi(E)$ is then defined as 
\[
    \Gamma(E) = \bigcup_{i, \ell} \Gamma_{i\ell}(E).
\]

\begin{theorem}\label{thm:main}
Any local minimum of the R$^3$-MSE \( \ell_{\lambda} \) is either: (1) a local minimum of the algebraic surrogate \( \ell_{\lambda,E} \) located in the partition interior \( \Psi(E)\setminus \Gamma(E) \), or (2) a local minimum of \( \ell_{\lambda,E} \) lying on the partition boundary $\Gamma(E)$, for some indicator matrix \( E \in \{ \pm 1 \}^{n \times L} \).
\end{theorem}

Proof of Theorem~\ref{thm:main} is provided in Appendix~\ref{app:proof_of_thm:main}. 

Therefore, we enumerate all local minima over both the interior and boundary, which guarantees that the resulting set includes all local minima of the R$^3$-MSE. However, since this set also contains some points that are not local minima in the unconstrained domain $\Psi$, we finally apply a filtering step to verify minimality and discard non-minimal points.

\section{Proposed Computational Algebraic Algorithm}
\label{sec:proposed_algorithm}
\label{ssec:3}

In this section, we present a computational algebraic algorithm for enumerating all local minima of the R$^3$-MSE~\eqref{eq:R3MSE} (which coincide with the local minima in RR-MSE~\eqref{eq:R2MSE}). 
Before providing a more rigorous description in Algorithm~\ref{alg:proposal}, we describe our strategy in Section~\ref{subsec:DEM} and an overview of the algorithm in Section~\ref{subsec:overview}. Subsequently, we explain more detailed algebraic operations in Section~\ref{subsec:detailed_algebraic_implementation}.

\subsection{Divide-Enumerate-Merge Strategy}
\label{subsec:DEM}

We first describe our strategy to construct the enumeration algorithm. 
A fundamental issue to consider at the outset is that, although the algebraic surrogate function is amenable to algebraic manipulation, its equivalence to the R$^3$-MSE holds only within the specific partition \( \Psi(E) \), and not over the entire parameter space \( \Psi \). Consequently, the application of computational algebra must be localized. We can only apply it meaningfully within each partition \( \Psi(E) \). 

This observation motivates our Divide-Enumerate-Merge strategy. The idea is to first divide the parameter space \( \Psi \) into partitions \( \{ \Psi(E) \} \) defined by all possible indicator matrices \( E \in \{\pm 1\}^{n \times L} \). For each partition $\Psi(E)$, we apply computational algebra to enumerate all local minima of the corresponding surrogate \( \ell_{\lambda,E}(\psi) \) (over the entire parameter space $\Psi$). We then filter out those local minima that fall outside the valid partition \( \Psi(E) \), as only those within \( \Psi(E) \) correspond to valid local minima of the R$^3$-MSE. Finally, we merge the valid local minima obtained from all indicator matrices \( E \), thereby recovering the complete set of local minima of the R$^3$-MSE in the entire parameter space $\Psi$. 
Therefore, it suffices to enumerate all local minima (candidates) within each partition \( \Psi(E) \), as detailed in the following section.

\subsection{Enumeration of Local Minimum Candidates in Each Partition: An Overview}
\label{subsec:overview}

This section provides an overview of how computational algebra serves as the foundation for enumerating all local minima within each partition. While this section focuses on describing the overview, a more detailed explanation of the algebraic implementation will be provided in Section~\ref{subsec:detailed_algebraic_implementation}.

Let \( \mathbb{R}[\psi] \) denote the polynomial ring, i.e., the set of polynomials in the variable \( \psi \), with real coefficients. In this study, we restrict our attention to real-valued polynomials, as complex-valued neural network parameters are beyond the scope of interest.  
Given polynomials \( f_1, f_2, \ldots, f_r \in \mathbb{R}[\psi] \), computational algebra offers a powerful framework for solving the corresponding system of equations $f_1(\psi)=0,f_2(\psi)=0,\ldots,f_r(\psi)=0$, whose solution set is known as an algebraic variety:
\begin{align}
    \mathbb{V}(f_1, f_2, \ldots, f_r)
    =
    \left\{
        \psi \in \Psi \ \middle| \ f_1(\psi) = 0,\, f_2(\psi) = 0,\, \ldots,\, f_r(\psi) = 0
    \right\}.
    \label{eq:algebraic_variety}
\end{align}

Therefore, to construct a complete list of local minima, algebraic varieties are systematically combined, with separate constructions tailored to the interior and boundary components of each partition $\Psi(E)$.

\subsubsection*{Interior of $\Psi(E)$} 

By definition, the algebraic surrogate \( \ell_{\lambda,E}(\psi) \) is represented as a ratio of polynomials. Since both its numerator and denominator are polynomial functions of the parameter \( \psi \), all partial derivatives of \( \ell_{\lambda,E}(\psi) \) with respect to \( \psi \) are also rational functions, each having polynomial numerators and denominators.

For any rational function \( r(\psi) \), we denote its numerator and denominator by \( \num(r(\psi)) \) and \( \den(r(\psi)) \), respectively. That is, $r(\psi) = \num(r(\psi))/\den(r(\psi))$. 
Since the equation \( r(\psi) = 0 \) is equivalent to \( \num(r(\psi)) = 0 \) and \( \den(r(\psi)) \neq 0 \), the set of internal stationary points (including all local minima) of the R$^3$-MSE, defined by the equation
\[
    \frac{\partial \ell_{\lambda,E}(\psi)}{\partial \psi} = 0,
\]
can be represented as the difference of two algebraic varieties:
\begin{align}
    \mathbb{V}(f_1, f_2, \ldots, f_w) \setminus \mathbb{V}\big(\prod_{j \le w+1} h_j\big).
    \label{eq:difference}
\end{align}
The polynomials $f_1,f_2,\ldots,f_w,h_1,h_2,\ldots,h_w,h_{w+1} \in \mathbb{R}[\psi]$ are defined as
\begin{align}
    f_1(\psi) &= \num \left( \frac{\partial \ell_{\lambda,E}(\psi)}{\partial b_{11}} \right), &
    f_2(\psi) &= \num \left( \frac{\partial \ell_{\lambda,E}(\psi)}{\partial b_{12}} \right), &
    \cdots \quad
    f_w(\psi) &= \num \left( \frac{\partial \ell_{\lambda,E}(\psi)}{\partial c_L} \right), \nonumber \\
    h_1(\psi) &= \den \left( \frac{\partial \ell_{\lambda,E}(\psi)}{\partial b_{11}} \right), &
    h_2(\psi) &= \den \left( \frac{\partial \ell_{\lambda,E}(\psi)}{\partial b_{12}} \right), &
    \cdots \quad
    h_w(\psi) &= \den \left( \frac{\partial \ell_{\lambda,E}(\psi)}{\partial c_L} \right), \nonumber \\
    &\text{and } \quad h_{w+1}(\psi)=\prod_{i,\ell}\xi_{i\ell}(\psi). 
    \label{eq:polynomials_interior}
\end{align}
The actual scale of the algebraic formulae~\eqref{eq:polynomials_interior} involved in our numerical experiments can be found in Supplement~\ref{supp:empirical_expression_scaling}.

Here, each member $\psi$ belonging to the set \eqref{eq:difference} satisfies the conditions 
$f_1(\psi)=0,f_2(\psi)=0,\ldots,f_w(\psi)=0,h_1(\psi)\ne 0,h_2(\psi) \ne 0,\ldots,h_w(\psi) \ne 0,h_{w+1}(\psi) \ne 0$. The final condition $h_{w+1}(\psi) \ne 0$, i.e., $\xi_{i\ell}(\psi) \ne 0$ for all $i$ and $\ell$, plays a crucial role in excluding points on the boundary $\Gamma(E)$ from being incorrectly identified as stationary points in the interior.

Such a difference between two algebraic varieties~\eqref{eq:difference} can be effectively handled using a technique called saturation, which will be introduced in Section~\ref{subsec:detailed_algebraic_implementation}.
Accordingly, computational algebra can be employed to enumerate all stationary points (including all local minima) of \( \ell_{\lambda,E}(\psi) \) over the entire parameter space \( \Psi = \mathbb{R}^{L \times d} \times \mathbb{R}^L \). 
We then discard any points that lie outside the partition \( \Psi(E) \) and retain only those in its interior. In this manner, we obtain a complete set of candidate local minima within the interior of the partition $\Psi(E)$.

\subsubsection*{Boundary of $\Psi(E)$} 
Theorem~\ref{thm:main} establishes that any local minimum of the R$^3$-MSE must also be a local minimum of the corresponding algebraic surrogate. Moreover, such minima may lie on the boundary, defined as \( \Gamma(E) = \bigcup_{i,\ell} \Gamma_{i\ell}(E) \). Accordingly, our implementation in this part systematically enumerates all minima located on each boundary component \( \Gamma_{i\ell}(E) \). By virtue of Theorem~\ref{thm:main}, each enumerated minimum qualifies as a valid candidate for a local minimum of the R$^3$-MSE.

Let \( (i,\ell) \in \{1, 2, \ldots, n\} \times \{1,2,\ldots,L\} \).  
To identify local minima within the boundary component \( \Gamma_{i \ell}(E) \), we consider the Lagrangian function \( 
L_{i\ell}(\psi,\beta)=\ell_{\lambda,E}(\psi) + \beta \, \xi_{i \ell}(\psi) \), where \( \beta \in \mathbb{R} \) denotes the Lagrange multiplier. Similarly to the interior of $\Psi(E)$, we solve the equation $\partial L_{i\ell}(\psi,\beta)/\partial \psi=0$ with the constraint $\xi_{i\ell}(\psi)=0$. The solution set is expressed as
\begin{align}
    \mathbb{V}(\widetilde{f}_1,\widetilde{f}_2,\ldots,\widetilde{f}_w,\widetilde{f}_{w+1})
    \setminus 
    \mathbb{V}\big(\prod_{j \le w} \widetilde{h}_j\big),
    \label{eq:difference-2}
\end{align}
where the polynomials $\widetilde{f}_1,\widetilde{f}_2,\ldots,\widetilde{f}_w,\widetilde{f}_{w+1},\widetilde{h}_1,\widetilde{h}_2,\ldots,\widetilde{h}_w \in \mathbb{R}[(\psi,\beta)]$ are defined as: 
\begin{align}
    \widetilde{f}_1(\psi,\beta) &= 
    \num\left(\frac{\partial L_{i\ell}(\psi,\beta)}{\partial b_{11}}\right), \quad 
    \widetilde{f}_2(\psi,\beta) = \num\left(\frac{\partial L_{i\ell}(\psi,\beta)}{\partial b_{12}}\right), \quad \cdots \quad 
    \widetilde{f}_w(\psi,\beta) = \num\left(\frac{\partial L_{i\ell}(\psi,\beta)}{\partial c_{L}}\right), \nonumber \\
    \widetilde{f}_{w+1}(\psi,\beta) &= \xi_{i \ell}(\psi), \nonumber \\
    \widetilde{h}_1(\psi,\beta) &= 
    \den\left(\frac{\partial L_{i\ell}(\psi,\beta)}{\partial b_{11}}\right), \quad 
    \widetilde{h}_2(\psi,\beta) = \den\left(\frac{\partial L_{i\ell}(\psi,\beta)}{\partial b_{12}}\right), \quad \cdots \quad 
    \widetilde{h}_w(\psi,\beta) = \den\left(\frac{\partial L_{i\ell}(\psi,\beta)}{\partial c_{L}}\right). 
    \label{eq:polynomials_boundary}
\end{align}
To focus solely on the essential parameter $\psi$ as in the interior case, the Lagrange multiplier $\beta$ is removed through an algebraic operation called elimination. Additionally, the difference between two algebraic varieties in~\eqref{eq:difference-2} can be effectively handled using the technique of saturation. Both of these algebraic techniques, i.e., elimination and saturation, will be introduced in Section~\ref{subsec:detailed_algebraic_implementation}.

Therefore, computational algebra can be employed to enumerate all local minima candidates of \( \ell_{\lambda,E}(\psi) \), even over the boundary component \( \Gamma_{i\ell} \). 
According to standard results in constrained optimization theory, the solutions obtained via the Lagrange multiplier method correspond to local extrema of R$^3$-MSE subject to the equality constraint $\xi_{i\ell}(\psi)=0$~(see, for instance, \citet{bertsekas1982constrained}; every pair of a local extremum $\psi^*$ and its corresponding Lagrange multiplier $\beta^*$ belongs to \eqref{eq:difference-2}).

\subsubsection*{Synthesis of Interior and Boundary Treatments}

We perform the above computation for each $ \in \{\pm 1\}^{n \times L}$, and consolidate the enumerated local minima across all partitions as discussed in Divide-Enumerate-Merge strategy. 
Although the enumerated points obtained from both the interior and boundary are local minima candidates, they are not guaranteed to be local minima of the R$^3$-MSE. Since it is straightforward to verify whether a given candidate is a valid local minimum of the R$^3$-MSE, we can conclude that the full set of valid local minima can be successfully enumerated.

\subsection{More Detailed Algebraic Operations}
\label{subsec:detailed_algebraic_implementation}

Having outlined the overall strategy and provided a general overview, we now delve into the algebraic implementation of our proposed algorithm in greater detail. 
To ensure accessibility for readers who may not be familiar with algebraic techniques, we carefully introduce and define the minimal set of essential concepts and notations as needed throughout the exposition. Additional details and foundational definitions are provided in Supplement~\ref{supp:basic_concepts}.

Firstly, as explained in Section~\ref{subsec:overview}, given polynomials \( f_1, f_2, \ldots, f_r \in \mathbb{R}[\psi] \), computational algebra can be used to compute the algebraic variety \( \mathbb{V}(f_1, f_2, \ldots, f_r) \) defined in~\eqref{eq:algebraic_variety}. 
Ideal is an important algebraic structure: $\mathcal{I} \subset \mathbb{R}[\psi]$ is called ideal if it satisfies (1) $0 \in \mathcal{I}$, (2) $f,f' \in \mathcal{I}\Rightarrow f+f' \in \mathcal{I}$, and 
(3) $\alpha \in \mathbb{R}[\psi],f \in \mathcal{I} \Rightarrow \alpha f \in \mathcal{I}$. 
We then define a polynomial ideal generated by \( f_1, f_2, \ldots, f_r \) as
\[
    \mathcal{I}
    =
    \ideal{f_1, f_2, \ldots, f_r}
    =
    \left\{
        \sum_{i=1}^r h_i(\psi) f_i(\psi) \,\, \bigg| \,\, h_i \in \mathbb{R}[\psi]
    \right\},
\]
and conversely, any ideal has a finite generating set, as stated in the following theorem.

\begin{theorem}[Hilbert Basis Theorem]
\label{thm:hilbert}
    Let \( \mathcal{I} \) be an ideal in the polynomial ring \( \mathbb{R}[\psi] \). 
    Then, \( \mathcal{I} \) has a finite generating set; that is, there exist 
    \( \widetilde{f}_1, \widetilde{f}_2, \ldots, \widetilde{f}_{\tilde{r}} \in \mathbb{R}[\psi] \) 
    such that $\mathcal{I} = \ideal{ \widetilde{f}_1, \widetilde{f}_2, \ldots, \widetilde{f}_{\tilde{r}} }$. 
\end{theorem}

For more details, see Theorem~4 in Section 2.5 of \citet{CLO}.
With this notation, the algebraic variety can alternatively be expressed as
\[
    \mathbb{V}(\mathcal{I})
    =
    \{ \psi \in \Psi \mid f(\psi) = 0,\ \forall f \in \mathcal{I} \} \quad (=\mathbb{V}(f_1,f_2,\ldots,f_r)).
\]

Hereinafter, let \( w = L(d+1) \) denote the number of entries in the parameter vector \( \psi = (\psi_1, \psi_2, \ldots, \psi_w) \in \Psi \).  
Let \( \mathbb{Z} \) denote the set of integers, and let \( a = (a_1, a_2, \ldots, a_w) \in \mathbb{Z}_{\ge 0}^w \) be a multi-index, referred to as a monomial exponent vector.  
Using the monomial \( \psi^a = \psi_1^{a_1} \psi_2^{a_2} \cdots \psi_w^{a_w} \), any polynomial \( f \in \mathbb{R}[\psi] \) can be expressed in the form
\[
  f(\psi) = \sum_{a \in \mathbb{Z}_{\ge 0}^w} c_a \psi^a, 
\]
where each \( c_a \in \mathbb{R} \) is a real coefficient indexed by $a$. 
To further analyze and manipulate algebraic varieties, we introduce three fundamental concepts from computational algebra: (1) Gr{\"o}bner basis, (2) saturation, and (3) elimination.

\subsubsection*{(1) Gr{\"o}bner basis}

Roughly speaking, in our context, a Gr{\"o}bner basis is a well-structured set of polynomials that generates a given ideal \( \mathcal{I} \). By effectively utilizing a Gr{\"o}bner basis, one can simplify the system of equations and decompose the polynomials to derive the solutions. In this subsection, we provide a formal definition of a Gr{\"o}bner basis.

We begin by introducing the concept of monomial orderings.
While ordering single scalar values is straightforward, defining a total order on vectors of multiple values, such as monomial exponent vectors $a=(a_1,a_2,\ldots,a_w)$, is significantly more challenging. In particular, there is no universally canonical way to order such vectors. As a result, various rules have been proposed to determine the order in which monomials should be arranged when representing and manipulating polynomials. Commonly used monomial orderings include the lexicographic order (often abbreviated as lex) and the graded reverse lexicographic order (often abbreviated as grevlex); see, for example, Section 2.2 of \citet{CLO}.

For each polynomial \( f \), the monomials can be ordered according to a specified monomial order.  
The monomial that appears first under this order is called the leading monomial, denoted by \( \mathrm{LM}(f) \). 
The leading monomial for the ideal $\mathcal{I}$ is also defined as $\mathrm{LM}(\mathcal{I})=\{c\psi^{a} \mid \exists f \in \mathcal{I} \setminus \{0\} \,\, \text{such that} \,\, \mathrm{LM}(f)=c\psi^a\}$. Then, the Gr{\"o}bner basis is defined as follows. 

\begin{definition}
Fix a monomial ordering on \( \mathbb{R}[\psi] \).  
A Gr{\"o}bner basis of an ideal \( \mathcal{I} \subset \mathbb{R}[\psi]\) is a finite subset \( G = \{g_1, g_2, \ldots, g_t\} \subset \mathcal{I}\) such that
\[
    \ideal{\mathrm{LM}(g_1), \mathrm{LM}(g_2), \ldots, \mathrm{LM}(g_t)}
    =
    \ideal{\mathrm{LM}(\mathcal{I})},
\]
where the right-hand side denotes the ideal generated by the leading monomials of all polynomials in \( \mathcal{I} \).
\end{definition}

Note that every polynomial ideal has a Gr{\"o}bner basis (see Corollary 6 in Section 2.5 of \citet{CLO}), and that any Gr{\"o}bner basis \( G \) of the ideal \( \mathcal{I} \) generates the same ideal, implying that the corresponding algebraic varieties are identical:
\[
\mathbb{V}(\mathcal{I}) = \mathbb{V}(G).
\]
When we use the lex order, Gr{\"o}bner bases can be viewed as simplified and lexicographically ordered generators of the ideal, where each polynomial incrementally eliminates variables one at a time. As a result, the variety \( \mathbb{V}(G) \) is often more amenable to computational methods, much like Gaussian elimination in linear algebra, than the original system \( \mathbb{V}(\mathcal{I}) \).

While Buchberger's algorithm provides the theoretical foundation for constructing Gr{\"o}bner bases~\citep{buchberger1976theoretical,CLO} from the given ideal $\mathcal{I}$, computing Gr{\"o}bner bases is generally known to be computationally intensive, with doubly exponential worst-case complexity in the number of variables~\citep{mayr1982complexity}. To address this, several algorithms have been developed to compute Gr{\"o}bner bases more efficiently. For instance, Faug{\`e}re's F4 and F5 algorithms~\citep{faugere1999F4,faugere2002F5} and M4GB~\citep{makarim2017M4GB}. 
More recently, \citet{kera2024learning} proposed an approach that leverages Transformer architectures in neural networks to approximate Gr{\"o}bner bases at significantly higher speeds, albeit with a loss of complete mathematical rigor.

Among the available software packages, several tools are commonly employed in practical implementations. Notable examples include \verb|Magma|~\citep{bosma1997magma}, \verb|Maple|~\citep{michael2005maple}, and \verb|Mathematica|~\citep{Mathematica}, all of which offer efficient capabilities for computing Gr{\"o}bner bases. In this study, we primarily use \verb|Magma|~\citep{bosma1997magma} to compute the Gr{\"o}bner bases of the ideals constructed in our framework.

\begin{remark}
    In general, Gr{\"o}bner bases with respect to the graded reverse lexicographic (grevlex) order can be computed more efficiently than those with respect to the lexicographic (lex) order. Moreover, if the given ideal is zero-dimensional, the FGLM algorithm~\citep{FGLM} enables efficient conversion of a Gr{\"o}bner basis from one monomial order to another. Accordingly, our algorithm first computes a Gr{\"o}bner basis with respect to the grevlex order, and then applies the FGLM algorithm to convert it to a lex order Gr{\"o}bner basis. 
\end{remark}

\subsubsection*{(2) Saturation ideal}
Roughly speaking, in our context, saturation is used to transform the ideal $\mathcal{I}$ into a more algebraically tractable form; it can be used to compute the the set-theoretic difference between algebraic varieties, as given in equations~\eqref{eq:difference} and~\eqref{eq:difference-2}.

Let \( \mathcal{I}, \mathcal{K} \) be ideals in the polynomial ring \( \mathbb{R}[\psi] \). The saturation of \( \mathcal{I} \) with respect to \( \mathcal{K} \) is defined as
\[
    \mathcal{I} : \mathcal{K}^{\infty}
    =
    \left\{
        f \in \mathbb{R}[\psi] \mid 
        \forall h \in \mathcal{K},\ \exists N \geq 0\ \text{such that } f h^N \in \mathcal{I}
    \right\}.
\]
Here, \( h^N \) denotes the \( N \)-th power of \( h \), and \( f h^N \) is the product of \( f \) and \( h^N \). 
The saturation \( \mathcal{I} : \mathcal{K}^{\infty} \) is itself an ideal (see Theorem 10 in Section 4.4 of \citet{CLO}), and we refer to it as the saturation ideal. By Theorem~\ref{thm:hilbert}, it also admits a finite generating set. Such a generating set for the saturation ideal $\mathcal{I}:\mathcal{K}^{\infty}$, i.e., Gr{\"o}bner basis, is obtained as follows. 

\begin{proposition}
Let \( \mathcal{I} = \ideal{f_1, f_2, \ldots, f_r} \) be an ideal in the polynomial ring \( \mathbb{R}[\psi] \), and let \( \mathcal{K} = \ideal{h} \) for some \( h \in \mathbb{R}[\psi] \).  
Define the extended ideal \( \widetilde{\mathcal{I}} = \ideal{f_1, f_2, \ldots, f_r, 1 - Yh} \subseteq \mathbb{R}[(\psi, Y)] \), where \( Y \) is a newly introduced variable.  
Let \( \widetilde{G} \) be a Gr{\"o}bner basis of \( \widetilde{\mathcal{I}} \) with respect to the lexicographic order.  
Then, the set \( G = \widetilde{G} \cap \mathbb{R}[\psi] \) is a Gr{\"o}bner basis for the saturation ideal \( \mathcal{I} : \mathcal{K}^{\infty} \).
\end{proposition}

The following equation also holds:
\begin{align}
    \mathbb{V}(\mathcal{I})
    \setminus 
    \mathbb{V}(\mathcal{K})
    =
    \mathbb{V}(\mathcal{I}:\mathcal{K}^{\infty})
    \setminus 
    \mathbb{V}(\mathcal{K}). 
    \label{eq:saturation_identity}
\end{align}
This is because the complex algebraic variety of the saturation ideal $\mathcal{I}:\mathcal{K}^{\infty}$ coincides with the Zariski closure of the set-theoretic difference between the complex algebraic variety of the ideal $\mathcal{I}$ and that of the ideal $\mathcal{K}$ (see Theorem 10 in Section 4.4 of \citet{CLO}); the same holds when restricted to real algebraic varieties, and thus \eqref{eq:saturation_identity} remains valid in our setting. 
Although computing a saturation ideal $\mathcal{I}:\mathcal{K}^{\infty}$ incurs additional computational cost, 
it often results in a simpler solution space compared to the original ideal $\mathcal{I}$, 
which in turn reduces the overall algebraic complexity of subsequent computations. 
Accordingly, our algorithm works with the saturation ideal $\mathcal{I}:\mathcal{K}^{\infty}$, 
rather than directly with the original ideal $\mathcal{I}$.

While taking the set-theoretic difference between two algebraic varieties \( V_1=\mathbb{V}(\mathcal{I}:\mathcal{K}^{\infty}) \) and \( V_2=\mathbb{V}(\mathcal{K}) \) may appear difficult, it becomes tractable once we obtain a Gr{\"o}bner basis \( G_1 \) such that \( V_1 = \mathbb{V}(G_1) \). In that case, all elements \( \psi \in V_1 \) can be enumerated algebraically, and the set-theoretic difference \( V_1 \setminus V_2 \) can be identified by filtering out those elements that also belong to \( V_2 \).

Consequently, by computing the set-theoretic difference \( V_1 \setminus V_2 \) corresponding to the ideals \( \mathcal{I} = \langle f_1, f_2, \ldots, f_w \rangle \) and \( \mathcal{K} = \langle h_1, h_2, \ldots, h_w, h_{w+1} \rangle \), where the polynomials are defined as in equation~\eqref{eq:polynomials_interior}, we can enumerate the entries in the target set~\eqref{eq:difference}. The set~\eqref{eq:difference-2} can also be computed using the same procedure.

\subsubsection*{(3) Elimination ideal}

Roughly speaking, elimination allows us to eliminate specific variables from an ideal, thereby yielding a new ideal in a smaller polynomial ring that captures the projection of the original variety onto a lower-dimensional coordinate space. 

Let \( \mathcal{I} \) be an ideal in the polynomial ring \( \mathbb{R}[\psi] \). 
For a positive integer $2 \le i \le r$, the \( i \)th elimination of \( \mathcal{I} \) is defined as
\[
    \elim{\mathcal{I}}_i 
    =
    \mathcal{I} \cap \mathbb{R}[(\psi_{1}, \psi_{2}, \ldots, \psi_{i-1})].
\]
As described in Exercise~1 of Section~3.1 in \citet{CLO}, elimination ideals are themselves ideals in the polynomial ring \( \mathbb{R}[\psi] \); hereafter, we refer to \( \elim{\mathcal{I}}_i \) as an elimination ideal. 
By Theorem~\ref{thm:hilbert}, each elimination ideal \( \elim{\mathcal{I}}_i \) has a finite generating set. Such a generating set for the $i$th elimination ideal $\elim{\mathcal{I}}_i$, i.e., Gr{\"o}bner basis, is obtained as follows.

\begin{proposition}
Let \( \mathcal{I} \) be an ideal in the polynomial ring \( \mathbb{R}[\psi] \).  
Let \( G \) be a Gr{\"o}bner basis of \( \mathcal{I} \) with respect to the lexicographic order.  
Then, the set \( G \cap \mathbb{R}[(\psi_{1}, \psi_{2}, \ldots, \psi_{i-1})] \) is a Gr{\"o}bner basis for the \( i \)th elimination ideal \( \elim{\mathcal{I}}_i \).
\end{proposition}

Recall that the system of algebraic equations includes a Lagrange multiplier. By applying elimination, we can remove the Lagrange multiplier and obtain a reduced system of equations defined solely in terms of the original parameters.

\subsubsection*{Overall algorithm}

By leveraging these concepts, we propose Algorithm~\ref{alg:proposal}. 
An illustrative example, which applies the proposed Algorithm~\ref{alg:proposal} to a simple setting, is provided in Supplement~\ref{supp:illustrative_example}.

\begin{algorithm}[!p]
\caption{Computing the complete set of local minima}
    \label{alg:proposal}
    \begin{algorithmic}[1]    
    \Require Fixed observations \(\{ (x_i, y_i) \}_{i=1}^n\), a hyper parameter \(\lambda>0\).
    \Ensure Candidates of all local minima in R$^3$-MSE \(\ell_{\lambda}(\psi)\).
    \State \(\psi \leftarrow (B,c)\): concatenation of the parameters of interest $B$ and $c$.
    \State \(S, S_{\partial}\) \(\leftarrow\) \(\emptyset\): variables to store the solutions in the interior and the boundary, respectively.
    \For{\(E = (e_{i\ell}) \in \{\pm 1\}^{n \times L}\)} 
    	\State \(\Psi(E) \leftarrow
        \left\{ \psi \in \Psi \mid \xi_{i\ell}(\psi) e_{i\ell} \ge 0, \, \forall i, \ell \right\} \): the target partition in the current iteration.
        \State \(\cov_{iE} \leftarrow \left(\frac{1}{2}(e_{i\ell}+1) \xi_{i\ell}(\psi)\right)_{\ell} \) for \(i = 1, \, 2, \, \ldots, n\).
        \State \(\COV_E \leftarrow (\cov_{1E}^{\top}, \cov_{2E}^{\top}, \ldots, \cov_{nE}^{\top})^{\top} \). 
        \State \(\ell_{\lambda,E}(\psi) \leftarrow\) the algebraic surrogate on $\Psi(E)$ defined in~\eqref{eq:algebraic_surrogate}.
        \State \(\Phi(E) \leftarrow \Psi(E) \cap \left\{ \psi \in \Psi \mid \xi_{i\ell}(\psi) \ne 0, \forall i,\ell \right\} \): the interior of \(\Psi(E)\).
        \State Ideals
        \[
           \mathcal{I} = \ideal{f_1,f_2,\ldots,f_w},
           ~~
           \mathcal{K} = \ideal{\prod_{j \le w+1}h_j}
        \]
        are generated from the polynomials $f_1,f_2,\ldots,f_w$ and $h_1,h_2,\ldots,h_w,h_{w+1}$ defined in \eqref{eq:polynomials_interior}
        \State \(G \leftarrow\) a Gr\"obner basis of the saturation \(\mathcal{I}:\mathcal{K}^{\infty}\) with respect to the lexicographic order.
        \Statex
        \Statex \quad \# \textbf{Processing the interior region}
        \State \(S \leftarrow S \cup ((\mathbb{V}(G) \setminus \mathbb{V}(\mathcal{K})) \cap \Phi(E))\).
        \Statex
        \Statex \quad \# \textbf{Processing the boundary components}
        \For{\((i,\ell) \in \{1,2,\ldots,n\} \times \{1,2,\ldots,L\}\)} 
            \State \(\Gamma_{i\ell}(E) \leftarrow \{\psi \in \Psi(E) \mid \xi_{i\ell}(\psi)=0 \} \): boundary component. 
            \State \(L_{\lambda,E}(\psi,\beta) \leftarrow \ell_{\lambda,E}(\psi) + \beta \xi_{i\ell}(\psi)\): Lagrange function, which belongs to $\mathbb{R}[(\psi,\beta)]$. 
            \State Ideals 
            \[
               \mathcal{I}_{i\ell} = \ideal{\widetilde{f}_1, \widetilde{f}_2, \ldots, \widetilde{f}_w, \widetilde{f}_{w+1}},
               ~~
               \mathcal{K}_{i\ell} = \ideal{\prod_{j \le w} \widetilde{h}_j}
            \]
            are generated from the polynomials $\widetilde{f}_1,\widetilde{f}_2,\ldots,\widetilde{f}_w,\widetilde{f}_{w+1}$ and $\widetilde{h}_1,\widetilde{h}_2,\ldots,\widetilde{h}_w$ defined in \eqref{eq:polynomials_boundary}.
            \State \(G_{i,\ell} \leftarrow\) a Gr{\"o}bner basis of the elimination (regarding the Lagrange multiplier $\beta$) of the saturation ideal \( \elim{\mathcal{I}_{i\ell}:\mathcal{K}_{i\ell}^{\infty}}_{w+1}\) with respect to the lexicographic order. 
            \State \(S_{\partial} \leftarrow S_{\partial} \cup \left((\mathbb{V}(G_{i\ell}) \setminus \mathbb{V}(\mathcal{K}_{i\ell})) \cap \Gamma_{i\ell}(E) \right)\).
        \EndFor 
    \EndFor 
    \State \(\Psi^{\dagger} \leftarrow S \cup S_{\partial}\).
    \State \(\Theta^{\dagger} \leftarrow \) reversed variable projection applied to $\Psi^{\dagger}$.
    \State \(\Theta^* \leftarrow\) discarding non-minimizing stationary points in \(\Theta^{\dagger}\).
    \State \Return \(\Theta^*\). 
    \end{algorithmic}
\end{algorithm}

\section{Numerical Visualization}
\label{sec:numerical}

In this section, we conduct numerical experiments to demonstrate the proposed algebraic algorithm. Specifically, we exhaustively enumerate all local minima in a minimal ReLU perceptron with a few hidden units.

\subsection{Implementation Details for Algebraic Computation}
\label{subsec:implementation_details}

\paragraph{Computation of Gr{\"o}bner basis:} 

In this study, we use \verb|Magma|~\citep{bosma1997magma} to compute Gr{\"o}bner bases, where the default setting employs the Monte-Carlo global-modular algorithm. This algorithm enables fast computation of Gr{\"o}bner bases at the cost of introducing errors with a probability on the order of $10^{-28}$. For details, see \url{https://magma.maths.usyd.edu.au/magma/handbook/text/1262#14427}.

\paragraph{Enumeration of non-zero dimensional solutions:} Although Algorithm~\ref{alg:proposal} preserves all the local minima, the set \( \Theta \) may contain not only zero-dimensional points (i.e., isolated stationary points including local minima), but also infinitely-many points forming higher-dimensional structures (i.e., curves, surfaces, or hypersurfaces), resulting in infinitely many candidates. 
In such cases, where the solution set includes higher-dimensional components, our numerical results visualize the structure by uniformly subsampling 10,000 points from within these components, rather than attempting to depict the entire continuous set.

\paragraph{Verification of local minimality:} 
If one seeks to rigorously verify local minimality, it is in principle possible to do so in a finite number of steps under the current setting, since the number of partitions is finite. Specifically, one can analytically check minimality along a suitable finite set of directions. However, due to the complexity of such an implementation, we opted for a more practical yet sufficiently reliable approach in this study. 
To verify whether each enumerated candidate point \( \psi \in \Psi \) is indeed a local minimum, we generate \( M = 20 \) randomly perturbed vectors \( \psi^{\dagger}_1, \psi^{\dagger}_2, \ldots, \psi^{\dagger}_M \) on a hypersphere of radius \( \varepsilon = 10^{-3} \) centered at \( \psi \), and check whether the inequality \( \ell_{\lambda}(\psi^{\dagger}_j) \geq \ell_{\lambda}(\psi) \) holds for all \( j = 1, 2, \ldots, M \). Although this procedure is a numerical verification and does not guarantee mathematical rigor in establishing optimality, it provides sufficient practical reliability in the simple problem setting considered in this study.

\subsection[Enumeration of Local Minima with L=2 Hidden Units]{Enumeration of Local Minima with \(L=2\) Hidden Units}
\label{subsec:results_L=2}

We employ an instance of the dataset (\( n=5,\, d=1 \)) defined as follows: 

\begin{align*}
(x_1,y_1) &= \left(-\frac{17}{100}, \frac{5}{100}\right), \quad 
(x_2,y_2) = \left( \frac{44}{100}, \frac{102}{100}\right), \quad 
(x_3,y_3) = \left(-\frac{100}{100}, \frac{61}{100} \right), \\ 
(x_4,y_4) &= \left(-\frac{40}{100}, -\frac{36}{100} \right), \quad 
(x_5,y_5) = \left(-\frac{71}{100}, -\frac{132}{100} \right). 
\end{align*}

\begin{figure}[!p]
\includegraphics[width=0.9\textwidth]{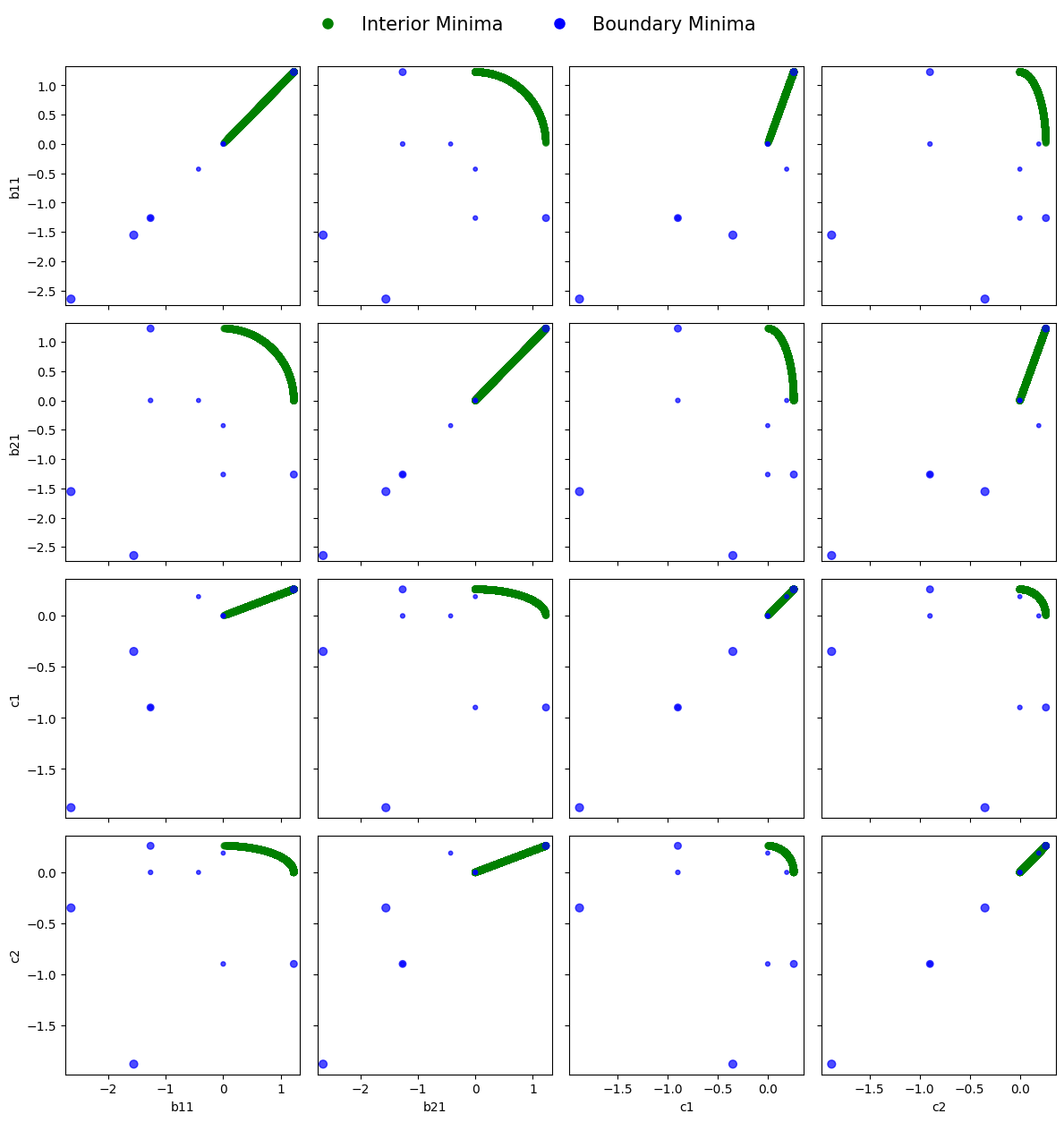}
\caption{Local minima for R$^3$-MSE, where $L=2,n=5$. Larger markers indicate smaller R$^3$-MSE values.}
\label{fig:L=2_n=5}
\end{figure}

For the case \(L = 2\), all local minima identified by our algorithm are visualized in Figure~\ref{fig:L=2_n=5}. In this setting, we compute \(2^{nL} = 2^{10} = 1024\) indicator matrices \(E = (e_{ij}) \in \{\pm 1\}^{n \times L}\), and for each \(E\), we solve the corresponding system of complicated algebraic equations. As a result, we detect $1$ (one-dimensional) connected set of local minima (i.e., a curve) in the interior, and $8$ (zero-dimensional) local minima (i.e., points) on the boundary. Detailed descriptions of these solutions are provided below.

\subsubsection*{Solutions in the Interior}

The one-dimensional connected set of local minima lies entirely within the interior region. This solution set is characterized as the collection of points \(\psi = (b_{11}, b_{21}, c_1, c_2)\) simultaneously satisfying all the constraints shown below: that is, inequalities
\begin{align*}
   c_1-\frac{17 b_{11}}{100}&>0, \quad
   c_2-\frac{17 b_{21}}{100}>0, \quad
   \frac{11 b_{11}}{25}+c_1>0, \quad
   \frac{11 b_{21}}{25}+c_2>0, \\
   c_1-B_{11} &<0, \quad
   c_2-B_{21} <0, \quad
   c_1-\frac{2 b_{11}}{5} <0, \quad
   c_2-\frac{2 b_{21}}{5} <0, \quad
   c_1-\frac{71 b_{11}}{100} <0 \quad
   c_2-\frac{71 b_{21}}{100} <0,
\end{align*}
and equations 
\begin{align*}
0 &=
b_{11}+
   R_1c_1^7 + R_2c_1^5c_2^2 + R_3c_1^5 
   + R_4c_1^3c_2^4 + R_5c_1^3c_2^2 + R_6c_1^3
   + R_7c_1c_2^6 + R_8c_1c_2^4 + R_9c_1c_2^2 - R_{10}c_1, \\
0 &=
b_{21} + R_{11}c_1^6c_2 + R_{12}c_1^4c_2^3 + R_{13}c_1^4c_2 
   + R_{14}c_1^2c_2^5 + R_{15}c_1^2c_2^3 + R_{16}c_1^2c_2
   + R_{17}c_2^7 + R_{18}c_2^5 + R_{19}c_2^3 - R_{20}c_2, \\
0 &= c_1^8 + 4c_1^6c_2^2 + R_{21}c_1^6 + 6c_1^4c_2^4 
    + R_{22}c_1^4c_2^2 + R_{23}c_1^4 + 4c_1^2c_2^6
    + R_{24}c_1^2c_2^4 + R_{25}c_1^2c_2^2 
    - R_{26}c_1^2 + c_2^8 \\
    &\hspace{5em} + R_{27}c_2^6 + R_{28} c_2^4
    - R_{29}c_2^2 - R_{30},
\end{align*}
where $R_1,R_2,\ldots,R_{30}$ are highly complicated rational numbers. These rational numbers are shown in Supplement~\ref{supp:coefficients}. 
We can easily verify that the set exhibits symmetry; specifically, if a pair \((b_{11}, c_1)\) and \((b_{21}, c_2)\) satisfies all the above inequalities and equations, then exchanging \((b_{11}, c_1)\) with \((b_{21}, c_2)\) yields another point that satisfies exactly the same conditions.

\subsubsection*{Solutions on the Boundary}

As an example, a certain element of~\eqref{eq:difference-2} is a solution of the following system of equations:
\begin{align*}
0 &= \text{univariate degree-32 polynomial in } c_{2}, \\
0 &= c_{1}^2 + (\text{polynomial in } c_{2}), \\
0 &= b_{21} + (\text{polynomial in } c_{1}, c_{2}), \\
0 &= b_{11} + (\text{polynomial in } b_{21}, c_{1}, c_{2}).
\end{align*}
In particular, \( c_{2} \) can be determined by solving the univariate degree-32 polynomial, which contains only even-degree terms. 
Since it is well known that no general closed-form solutions exist for polynomials of degree greater than four, 
representing the roots of such a degree-32 polynomial explicitly is highly challenging. 
Therefore, in this study, we employ a numerical solver; 
even a classical method such as Newton's method can compute numerical solutions for \( c_{2} \) with arbitrary precision. 
Once \( c_{2} \) is obtained, the corresponding \( c_{1} \) can be determined, 
followed by \( b_{21} \) and \( b_{11} \) in sequence.

With this calculation in mind, we can enumerate all the entries of~\eqref{eq:difference-2} and subsequently verify their minimality. 
As a result, we identify the 8 local minima on the boundary, which are listed in Table~\ref{table:minima_on_the_boundary}. 
Similar to the solutions in the interior, these boundary minima also exhibit symmetry: 
if \( (b_{11}, b_{21}, c_{1}, c_{2}) \) is a local minimum, then the permuted point \( (b_{21}, b_{11}, c_{2}, c_{1}) \) is also a local minimum.

\begin{table}[!ht]
\centering 
\caption{Detected 8 local minima on the boundary.}
\label{table:minima_on_the_boundary}
\begin{tabular}{c|rrrr}
\toprule 
ID & $b_{11}$ & $b_{21}$ & $c_1$ & $c_2$ \\
\hline
$1$ & $-2.645858$ & $-1.554645$ & $-1.878559$ & $-0.348159$ \\
$2$ & $-1.554645$ & $-2.645858$ & $-0.348159$ & $-1.878559$ \\
$3$ & $-1.263412$ & $1.227047$ & $-0.897023$ & $0.260054$ \\
$4$ & $1.227047$ & $-1.263412$ & $0.260054$ & $-0.897023$ \\
$5$ & $0$ & $-1.263412$ & $0$ & $-0.897023$ \\
$6$ & $-1.263412$ & $0$ & $-0.897023$ & $0$ \\
$7$ & $0$ & $-0.429247$ & $0$ & $0.188869$ \\
$8$ & $-0.429247$ & $0$ & $0.188869$ & $0$ \\
\bottomrule
\end{tabular}
\end{table}

\subsubsection*{Computational complexity}
Computational algebra has the advantage of being able to enumerate all algebraic solutions, 
in contrast to optimization algorithms such as gradient descent that typically converge only to a single local solution or stationary point. 
However, this advantage comes at the cost of extremely high computational complexity, 
making it difficult to apply the method to neural networks of practical size. 
In particular, for each indicator matrix $E \in \{\pm 1\}^{n \times L}$ and the corresponding region, 
one must compute a Gr{\"o}bner basis. 
Since the number of such regions grows exponentially as $2^{nL}$ with the sample size $n$, 
the computational burden increases rapidly. 
For instance, in the numerical visualization above, 
the parameter space is divided into $2^{5 \times 2} = 2^{10} = 1024$ regions, 
and a Gr{\"o}bner basis is computed for each. 
If the number of hidden units increases from $L=2$ to $L=3$, 
the number of regions explodes to $2^{5 \times 3} = 2^{15} = 32768$, 
and in addition, the number of parameters involved in the algebraic equations within each region also increases 
(see Supplement~\ref{supp:varying_width}), 
making the computation of a single Gr{\"o}bner basis itself prohibitively difficult. 
To the best of our knowledge, this problem setting, i.e., the need to compute Gr{\"o}bner bases for a massive number of 
parallel but slightly different algebraic systems, is not widely studied in the literature, 
and developing more efficient computational methods for such tasks remains an important challenge for future work.

\section{Conclusion}
\label{sec:conclusion}

In this paper, we presented a computational algebraic approach to rigorously analyze the loss landscape of a minimal ReLU perceptron with ridge regularization. By exploiting the piecewise polynomial structure of the loss function, we systematically enumerated all substationary points using an algebraic algorithm based on a Divide-Enumerate-Merge strategy.

\section*{Acknowledgement}

R. Fukasaku was supported by JSPS KAKENHI (23K10988, 23K03123, 24K10903, 24K03079, 25K03087, 25H01482). 
Y. Kabata was supported by JSPS KAKENHI (24K22308, 24K10903, 24K03079, 25K00208, 25H01485). 
A. Okuno was supported by JSPS KAKENHI (21K17718, 25K03087). 
We would like to thank Shotaro Yagishita for the helpful discussions.

\appendix

\section{Proofs}
\label{app:proofs}

\subsection[Derivation of R3-MSE]{Derivation of R$^3$-MSE}
\label{app:derivation_of_eq:R3MSE}

Substituting $\hat{a}=(\COV^{\top}\COV+\lambda I_L)^{-1}\COV^{\top}\ty$ into RR-MSE~\eqref{eq:R2MSE} yields
\begin{align}
    \min_{a \in \mathbb{R}^L} \widetilde{\ell}(a,B,c) 
&=
    \widetilde{\ell}(\hat{a},B,c) \nonumber \\
&=
\lambda \|\psi\|_2^2
+
\|\ty - \COV\hat{a}\|_2^2
+
\lambda \|\hat{a}\|_2^2 \bigg|_{\hat{a}=(\COV^{\top}\COV+\lambda I_L)^{-1}\COV^{\top}\ty} \nonumber \\
&=
\lambda \|\psi\|_2^2
+
\|\ty - \COV(\COV^{\top}\COV+\lambda I_L)^{-1}\COV^{\top}\ty\|_2^2
+
\lambda \|(\COV^{\top}\COV+\lambda I_L)^{-1}\COV^{\top}\ty\|_2^2 \nonumber \\
&=
\lambda \|\psi\|_2^2
+
\ip{\ty}{ 
\{I-\COV(\COV^{\top}\COV+\lambda I_L)^{-1}\COV^{\top}\}^2 \ty
}
+
\lambda 
\ip{\ty}{ 
\COV(\COV^{\top}\COV+\lambda I_L)^{-2}\COV^{\top} \ty
} \nonumber \\
&=
\lambda \|\psi\|_2^2
+
\ip{\ty}{ 
    [
        \underbrace{\{I-\COV(\COV^{\top}\COV+\lambda I_L)^{-1}\COV^{\top}\}^2}_{(\star)}
        +
        \lambda (\COV^{\top}\COV+\lambda I_L)^{-2}\COV^{\top}
    ]
    \ty
}. \label{eq:R3MSE-intermediate1}
\end{align}
As $(\star)$ can be expanded as
\begin{align*}
    (\star)
&=
    I
    -
    2\COV(\COV^{\top}\COV+\lambda I_L)^{-1}\COV^{\top}
    +
    \COV(\COV^{\top}\COV+\lambda I_L)^{-1}
    \underbrace{\COV^{\top}\COV}_{=\COV^{\top}\COV+\lambda I_L-\lambda I_L}
    (\COV^{\top}\COV+\lambda I_L)^{-1}\COV^{\top} \\
&=
    I
    -
    2\COV(\COV^{\top}\COV+\lambda I_L)^{-1}\COV^{\top}
    +
    \COV(\COV^{\top}\COV+\lambda I_L)^{-1}\COV^{\top}
    +
    \lambda \COV(\COV^{\top}\COV+\lambda I_L)^{-2}\COV^{\top}, 
\end{align*}
\eqref{eq:R3MSE-intermediate1} further reduces to:
\begin{align*}
    \eqref{eq:R3MSE-intermediate1}
    &=
    \lambda \|\psi\|_2^2
    +
    \ip{\ty}{
    \{I-\COV(\COV^{\top}\COV+\lambda I_L)^{-1}\COV^{\top}\}\ty
    }
    =
    \lambda \|\psi\|_2^2
    +
    \|\ty\|_2^2
    -
    \ip{\ty}{\COV(\COV^{\top}\COV+\lambda I_L)^{-1}\COV^{\top}\ty},
\end{align*}
which coincides with R$^3$-MSE~\eqref{eq:R3MSE} up to the constant term $\|\ty\|_2^2$. 
\qed

\subsection{Proof of Theorem~\ref{thm:main}}
\label{app:proof_of_thm:main}

This proof consists of two parts: the interior and the boundary. 

\paragraph{Interior:} Fix an indicator matrix \(E \in \{\pm 1\}^{n \times L}\). 
We define the interior of the partition \(\Psi(E)\) as
\[
    \Phi(E) = \Psi(E) \setminus \Gamma(E),
\]
and consider its minima and maxima. 
Since the algebraic surrogate \( \ell_{\lambda,E} \) coincides with the mean squared error \( \ell_{\lambda} \) over \( \Phi(E) \), and both functions are differentiable on this region, the following relation holds:
\begin{align}\label{eq:thm:main:proof}
    \left\{
      \psi \in \Phi(E) \mid \frac{\partial \ell_{\lambda,E}(\psi)}{\partial \psi} = 0
    \right\}
    =
    \left\{
      \psi \in \Phi(E) \mid \frac{\partial \ell_{\lambda}(\psi)}{\partial \psi} = 0
    \right\}.
\end{align}
That is, the set of stationary points of \( \ell_{\lambda,E} \) contained in \( \Phi(E) \) coincides with the set of stationary points of \( \ell_{\lambda} \) contained in \( \Phi(E) \).  
Therefore, all minima and maxima of \( \ell_{\lambda} \) within \( \Phi(E) \) are included in the left-hand side of the equation \eqref{eq:thm:main:proof}.

\paragraph{Boundary:} 
There exists an index \( i \in \{1,2, \ldots, n\}, \ell \in \{1,2,\ldots,L\} \) such that \( \psi \in \Gamma_i(E) \). 
Therefore, in what follows, we consider a minima located on a boundary component \( \Gamma_{i\ell}(E) \), and the same argument applies when $\psi$ is a maximum. 

Although these points may not be differentiable points of \( \ell_{\lambda} \), each of them can still be a minima.  
Recall that a point \( \psi \in \Gamma_{i\ell}(E) \) is a local minima of \( \ell_{\lambda} \) if and only if
\[
  \exists \varepsilon > 0,  \quad \forall \phi \in \Psi, \quad 
  \| \psi - \phi \| < \varepsilon \Rightarrow \ell_{\lambda}(\psi) \leq \ell_{\lambda}(\phi).
\]

Therefore, by the inclusion \(\Gamma_{i\ell}(E) \subset \Psi\) and the arbitrariness of \(\phi \in \Psi\), if \(\psi\) is a local minima of \(\ell_{\lambda}\) contained in \(\Gamma_{i\ell}(E)\), then we have one of the following:
\begin{align*}
  \exists \varepsilon > 0, \quad \forall \phi \in \Gamma_{i\ell}(E), \quad \|\psi - \phi\| < \varepsilon \Rightarrow \ell_{\lambda}(\psi) \leq  \ell_{\lambda}(\phi),
\end{align*}

As the algebraic surrogate \(\ell_{\lambda,E}\) is equivalent to \(\ell_{\lambda}\) for any \(\psi \in \Gamma_{i\ell}(E) \subset \Psi(E)\), we have 
\begin{align*}
  \exists \varepsilon > 0, \quad \forall \phi \in \Gamma_{i\ell}(E), \quad \|\psi - \phi\| < \varepsilon \Rightarrow \ell_{\lambda,E}(\psi) \leq  \ell_{\lambda,E}(\phi),
\end{align*}
indicating that \(\psi\) is also a local minima of the algebraic surrogate \(\ell_{\lambda,E}\) over the boundary component $\Gamma_{i\ell}(E)$. 

\qed

\bibliographystyle{apalike}
\bibliography{caann}

\begin{thebibliography}{}

\bibitem[Abramson et~al., 2024]{abramson2024accurate}
Abramson, J., Adler, J., Dunger, J., Evans, R., Green, T., Pritzel, A., Ronneberger, O., Willmore, L., Ballard, A.~J., Bambrick, J., et~al. (2024).
\newblock Accurate structure prediction of biomolecular interactions with alphafold 3.
\newblock {\em Nature}, 630:493--500.

\bibitem[Alfarra et~al., 2022]{alfarra2022decision}
Alfarra, M., Bibi, A., Hammoud, H., Gaafar, M., and Ghanem, B. (2022).
\newblock On the decision boundaries of neural networks: A tropical geometry perspective.
\newblock {\em IEEE Transactions on Pattern Analysis and Machine Intelligence}, 45(4):5027--5037.

\bibitem[Aoyagi, 2010]{aoyagi2010boltzmann}
Aoyagi, M. (2010).
\newblock Stochastic complexity and generalization error of a restricted boltzmann machine in bayesian estimation.
\newblock {\em Journal of Machine Learning Research}, 11(41):1243--1272.

\bibitem[Aoyagi, 2019]{aoyagi2019vandermonde}
Aoyagi, M. (2019).
\newblock Learning coefficient of vandermonde matrix-type singularities in model selection.
\newblock {\em Entropy}, 21(6).

\bibitem[Aoyagi, 2024]{aoyagi2024consideration}
Aoyagi, M. (2024).
\newblock Consideration on the learning efficiency of multiple-layered neural networks with linear units.
\newblock {\em Neural Networks}, 172:106132.

\bibitem[Aoyagi and Watanabe, 2005]{aoyagi2005stochastic}
Aoyagi, M. and Watanabe, S. (2005).
\newblock Stochastic complexities of reduced rank regression in bayesian estimation.
\newblock {\em Neural Networks}, 18(7):924--933.

\bibitem[Arena et~al., 1997]{arena1997quaternion}
Arena, P., Fortuna, L., Muscato, G., and Xibilia, M. (1997).
\newblock Multilayer perceptrons to approximate quaternion valued functions.
\newblock {\em Neural Networks}, 10(2):335--342.

\bibitem[Arora et~al., 2018]{arora2018understanding}
Arora, R., Basu, A., Mianjy, P., and Mukherjee, A. (2018).
\newblock Understanding deep neural networks with rectified linear units.
\newblock In {\em International Conference on Learning Representations}.

\bibitem[Bartlett et~al., 2017]{Bartlett2017Spectrally}
Bartlett, P.~L., Foster, D.~J., and Telgarsky, M. (2017).
\newblock Spectrally-normalized margin bounds for neural networks.
\newblock {\em Advances in Neural Information Processing Systems}, 30.

\bibitem[Bertsekas, 1982]{bertsekas1982constrained}
Bertsekas, D.~P. (1982).
\newblock {\em Constrained optimization and Lagrange multiplier methods}, volume Computer Science and Applied Mathematics of {\em Monographs and Textbooks}.
\newblock Academic press.

\bibitem[Bishop, 1995]{bishop1995neural}
Bishop, C.~M. (1995).
\newblock {\em Neural Networks for Pattern Recognition}.
\newblock Oxford University Press.

\bibitem[Boob et~al., 2022]{boob2022complexity}
Boob, D., Dey, S.~S., and Lan, G. (2022).
\newblock Complexity of training relu neural network.
\newblock {\em Discrete Optimization}, 44:100620.

\bibitem[Bosma et~al., 1997]{bosma1997magma}
Bosma, W., Cannon, J., and Playoust, C. (1997).
\newblock The {M}agma algebra system. {I}. {T}he user language.
\newblock {\em J. Symbolic Comput.}, 24(3-4):235--265.
\newblock Computational algebra and number theory (London, 1993).

\bibitem[Brandenburg et~al., 2024]{brandenburg2024the}
Brandenburg, M.-C., Loho, G., and Montufar, G. (2024).
\newblock The real tropical geometry of neural networks for binary classification.
\newblock {\em Transactions on Machine Learning Research}.

\bibitem[Brown et~al., 2020]{brown2020gpt}
Brown, T., Mann, B., Ryder, N., Subbiah, M., Kaplan, J.~D., Dhariwal, P., Neelakantan, A., Shyam, P., Sastry, G., Askell, A., Agarwal, S., Herbert-Voss, A., Krueger, G., Henighan, T., Child, R., Ramesh, A., Ziegler, D., Wu, J., Winter, C., Hesse, C., Chen, M., Sigler, E., Litwin, M., Gray, S., Chess, B., Clark, J., Berner, C., McCandlish, S., Radford, A., Sutskever, I., and Amodei, D. (2020).
\newblock Language models are few-shot learners.
\newblock In {\em Advances in Neural Information Processing Systems}, volume~33, pages 1877--1901. Curran Associates, Inc.

\bibitem[Buchberger, 1976]{buchberger1976theoretical}
Buchberger, B. (1976).
\newblock A theoretical basis for the reduction of polynomials to canonical forms.
\newblock {\em ACM SIGSAM Bulletin}, 10(3):19–29.

\bibitem[Buchholz and Sommer, 2001]{Buchholz2001}
Buchholz, S. and Sommer, G. (2001).
\newblock {\em Introduction to Neural Computation in Clifford Algebra}, pages 291--314.
\newblock Springer Berlin Heidelberg, Berlin, Heidelberg.

\bibitem[Buchholz and Sommer, 2008]{Buchholz2008clifford}
Buchholz, S. and Sommer, G. (2008).
\newblock On clifford neurons and clifford multi-layer perceptrons.
\newblock {\em Neural Networks}, 21(7):925--935.

\bibitem[Chang et~al., 2024]{chang2024llm}
Chang, Y., Wang, X., Wang, J., Wu, Y., Yang, L., Zhu, K., Chen, H., Yi, X., Wang, C., Wang, Y., Ye, W., Zhang, Y., Chang, Y., Yu, P.~S., Yang, Q., and Xie, X. (2024).
\newblock A survey on evaluation of large language models.
\newblock {\em ACM Transactions on Intelligent Systems and Technology}, 15(39):1--45.

\bibitem[Charisopoulos and Maragos, 2019]{charisopoulos2019tropical}
Charisopoulos, V. and Maragos, P. (2019).
\newblock A tropical approach to neural networks with piecewise linear activations.

\bibitem[Chen et~al., 2019]{chen2019regularized}
Chen, G.-Y., Gan, M., Chen, C. L.~P., and Li, H.-X. (2019).
\newblock A regularized variable projection algorithm for separable nonlinear least-squares problems.
\newblock {\em IEEE Transactions on Automatic Control}, 64(2):526--537.

\bibitem[Cox et~al., 2015]{CLO}
Cox, D.~A., Little, J., and O'Shea, D. (2015).
\newblock {\em Ideals, Varieties, and Algorithms: An Introduction to Computational Algebraic Geometry and Commutative Algebra}.
\newblock Springer Publishing Company, Incorporated, 4th edition.

\bibitem[Cruz et~al., 2018]{cruz2018neural}
Cruz, R.~S., Fernando, B., Cherian, A., and Gould, S. (2018).
\newblock Neural algebra of classifiers.
\newblock In {\em 2018 IEEE Winter Conference on Applications of Computer Vision}, pages 729--737.

\bibitem[Cybenko, 1989]{Cybenko1989}
Cybenko, G. (1989).
\newblock Approximation by superpositions of a sigmoidal function.
\newblock {\em Mathematics of Control, Signals and Systems}, 2(4):303--314.

\bibitem[Dauphin et~al., 2014]{dauphin2014identifying}
Dauphin, Y.~N., Pascanu, R., Gulcehre, C., Cho, K., Ganguli, S., and Bengio, Y. (2014).
\newblock Identifying and attacking the saddle point problem in high-dimensional non-convex optimization.
\newblock In {\em Advances in Neural Information Processing Systems}, volume~27. Curran Associates, Inc.

\bibitem[Dinh et~al., 2017]{dinh2017sharp}
Dinh, L., Pascanu, R., Bengio, S., and Bengio, Y. (2017).
\newblock Sharp minima can generalize for deep nets.
\newblock In {\em Proceedings of the 34th International Conference on Machine Learning}, volume~70 of {\em Proceedings of Machine Learning Research}, pages 1019--1028. PMLR.

\bibitem[Dosovitskiy et~al., 2021]{dosovitskiy2021an}
Dosovitskiy, A., Beyer, L., Kolesnikov, A., Weissenborn, D., Zhai, X., Unterthiner, T., Dehghani, M., Minderer, M., Heigold, G., Gelly, S., Uszkoreit, J., and Houlsby, N. (2021).
\newblock An image is worth 16x16 words: Transformers for image recognition at scale.
\newblock In {\em International Conference on Learning Representations}.

\bibitem[Drton and Plummer, 2017]{drton2017bayesian}
Drton, M. and Plummer, M. (2017).
\newblock A bayesian information criterion for singular models.
\newblock {\em Journal of the Royal Statistical Society Series B: Statistical Methodology}, 79(2):323--380.

\bibitem[Elbr{\"a}chter et~al., 2021]{elbrachter2021deep}
Elbr{\"a}chter, D., Perekrestenko, D., Grohs, P., and B{\"o}lcskei, H. (2021).
\newblock Deep neural network approximation theory.
\newblock {\em IEEE Transactions on Information Theory}, 67(5):2581--2623.

\bibitem[Español and Pasha, 2023]{espanol2023variable}
Español, M.~I. and Pasha, M. (2023).
\newblock Variable projection methods for separable nonlinear inverse problems with general-form tikhonov regularization.
\newblock {\em Inverse Problems}, 39(8):084002.

\bibitem[Faug\`{e}re, 1999]{faugere1999F4}
Faug\`{e}re, J.~C. (1999).
\newblock A new efficient algorithm for computing {G}robner bases (f4).
\newblock {\em Journal of Pure and Applied Algebra}, 139(1):61--88.

\bibitem[Faug\`{e}re, 2002]{faugere2002F5}
Faug\`{e}re, J.~C. (2002).
\newblock A new efficient algorithm for computing {G}r\"{o}bner bases without reduction to zero (f5).
\newblock In {\em Proceedings of the International Symposium on Symbolic and Algebraic Computation}, ISSAC '02, page 75–83, New York, NY, USA. Association for Computing Machinery.

\bibitem[Faug\`{e}re et~al., 1993]{FGLM}
Faug\`{e}re, J.~C., Gianni, P., Lazard, D., and Mora, T. (1993).
\newblock Efficient computation of zero-dimensional gr\"obner bases by change of ordering.
\newblock {\em Journal of Symbolic Computation}, 16(4):329--344.

\bibitem[Fiat et~al., 2019]{fiat2019decoupling}
Fiat, J., Malach, E., and Shalev-Shwartz, S. (2019).
\newblock Decoupling gating from linearity.

\bibitem[Fukasaku et~al., 2024]{fukasaku2024algebraic}
Fukasaku, R., Hirose, K., Kabata, Y., and Teramoto, K. (2024).
\newblock Algebraic approach to maximum likelihood factor analysis.
\newblock \url{https://arxiv.org/abs/2402.08181}.

\bibitem[Fukasaku et~al., 2025]{fukasaku2025algebraic}
Fukasaku, R., Yamamoto, M., Kabata, Y., Ikematsu, Y., and Hirose, K. (2025).
\newblock Algebraic approach for orthomax rotations.
\newblock \url{https://arxiv.org/abs/2504.21288}.

\bibitem[Fukushima, 1980]{fukushima1980neocognitron}
Fukushima, K. (1980).
\newblock Neocognitron: {A} self-organizing neural network model for a mechanism of pattern recognition unaffected by shift in position.
\newblock {\em Biological cybernetics}, 36(4):193--202.

\bibitem[Funahashi, 1989]{Funahashi1989}
Funahashi, K.-I. (1989).
\newblock On the approximate realization of continuous mappings by neural networks.
\newblock {\em Neural Networks}, 2(3):183--192.

\bibitem[Ghorbani et~al., 2019]{Ghorbani2019Hessian}
Ghorbani, B., Krishnan, S., and Xiao, Y. (2019).
\newblock An investigation into neural net optimization via hessian eigenvalue density.
\newblock In {\em Proceedings of the 36th International Conference on Machine Learning}, volume~97 of {\em Proceedings of Machine Learning Research}, pages 2232--2241. PMLR.

\bibitem[Golub and Pereyra, 2003]{golub2003separable}
Golub, G. and Pereyra, V. (2003).
\newblock Separable nonlinear least squares: the variable projection method and its applications.
\newblock {\em Inverse Problems}, 19(2):R1.

\bibitem[Golub and Pereyra, 1973]{golub1973differentiation}
Golub, G.~H. and Pereyra, V. (1973).
\newblock The differentiation of pseudo-inverses and nonlinear least squares problems whose variables separate.
\newblock {\em SIAM Journal on Numerical Analysis}, 10(2):413--432.

\bibitem[Goodfellow et~al., 2016]{Goodfellow-et-al-2016}
Goodfellow, I., Bengio, Y., and Courville, A. (2016).
\newblock {\em Deep Learning}.
\newblock MIT Press.
\newblock \url{http://www.deeplearningbook.org}.

\bibitem[Hashimoto et~al., 2022]{Hashimoto2021}
Hashimoto, Y., Wang, Z., and Matsui, T. (2022).
\newblock C*-algebra net: A new approach generalizing neural network parameters to c*-algebra.
\newblock In {\em Proceedings of the 39th International Conference on Machine Learning}, volume 162 of {\em Proceedings of Machine Learning Research}, pages 8523--8534. PMLR.

\bibitem[Hirose, 1992]{hirose1992proposal}
Hirose, A. (1992).
\newblock Proposal of fully complex-valued neural networks.
\newblock In {\em Proceedings of the International Joint Conference on Neural Networks}, volume~4, pages 152--157.

\bibitem[Hornik, 1991]{Hornik1989}
Hornik, K. (1991).
\newblock Approximation capabilities of multilayer feedforward networks.
\newblock {\em Neural Networks}, 4(2):251--257.

\bibitem[Jain et~al., 1996]{jain1996artificial}
Jain, A.~K., Mao, J., and Mohiuddin, K.~M. (1996).
\newblock Artificial {N}eural {N}etworks: {A} {T}utorial.
\newblock {\em Computer}, 29(3):31--44.

\bibitem[Jumper et~al., 2021]{jumper2021highly}
Jumper, J., Evans, R., Pritzel, A., Green, T., Figurnov, M., Ronneberger, O., Tunyasuvunakool, K., Bates, R., {\v{Z}}{\'\i}dek, A., Potapenko, A., et~al. (2021).
\newblock Highly accurate protein structure prediction with alphafold.
\newblock {\em Nature}, 596(7873):583--589.

\bibitem[Kawaguchi, 2016]{kawaguchi2016deep}
Kawaguchi, K. (2016).
\newblock Deep learning without poor local minima.
\newblock In {\em Advances in Neural Information Processing Systems}, volume~29. Curran Associates, Inc.

\bibitem[Kawaguchi et~al., 2019]{kawaguchi2019every}
Kawaguchi, K., Huang, J., and Kaelbling, L.~P. (2019).
\newblock Every local minimum value is the global minimum value of induced model in nonconvex machine learning.
\newblock {\em Neural Computation}, 31(12):2293--2323.

\bibitem[Kera et~al., 2024]{kera2024learning}
Kera, H., Ishihara, Y., Kambe, Y., Vaccon, T., and Yokoyama, K. (2024).
\newblock Learning to compute {G}r\"{o}bner bases.
\newblock In {\em Advances in Neural Information Processing Systems}, volume~37, pages 33141--33187. Curran Associates, Inc.

\bibitem[Keskar et~al., 2017]{Keskar2017LargeBatch}
Keskar, N.~S., Mudigere, D., Nocedal, J., Smelyanskiy, M., and Tang, P. T.~P. (2017).
\newblock On large-batch training for deep learning: Generalization gap and sharp minima.
\newblock In {\em International Conference on Learning Representations}.

\bibitem[Kim and Pilanci, 2024]{kim2024convex}
Kim, S. and Pilanci, M. (2024).
\newblock Convex relaxations of {R}e{LU} neural networks approximate global optima in polynomial time.
\newblock In {\em Proceedings of the 41st International Conference on Machine Learning}, volume 235 of {\em Proceedings of Machine Learning Research}, pages 24458--24485. PMLR.

\bibitem[Kipf and Welling, 2017]{kipf2017semisupervised}
Kipf, T.~N. and Welling, M. (2017).
\newblock Semi-supervised classification with graph convolutional networks.
\newblock In {\em International Conference on Learning Representations}.

\bibitem[Krizhevsky et~al., 2012]{alexnet}
Krizhevsky, A., Sutskever, I., and Hinton, G.~E. (2012).
\newblock Imagenet classification with deep convolutional neural networks.
\newblock In {\em Advances in Neural Information Processing Systems}, volume~25. Curran Associates, Inc.

\bibitem[Kurumadani, 2025]{kurumadani2025learning}
Kurumadani, Y. (2025).
\newblock Learning coefficients in semiregular models i: properties.
\newblock {\em Japanese Journal of Statistics and Data Science}, pages 1--29.

\bibitem[LeCun et~al., 2015]{LeCun2015}
LeCun, Y., Bengio, Y., and Hinton, G. (2015).
\newblock Deep learning.
\newblock {\em Nature}, 521:436--444.

\bibitem[LeCun et~al., 1989]{lecun1989backpropagation}
LeCun, Y., Boser, B., Denker, J.~S., Henderson, D., Howard, R.~E., Hubbard, W., and Jackel, L.~D. (1989).
\newblock Backpropagation applied to handwritten zip code recognition.
\newblock {\em Neural Computation}, 1(4):541--551.

\bibitem[Li et~al., 2018]{LiLossLandscape2018}
Li, H., Xu, Z., Taylor, G., Studer, C., and Goldstein, T. (2018).
\newblock Visualizing the loss landscape of neural nets.
\newblock In {\em Advances in Neural Information Processing Systems}, volume~31. Curran Associates, Inc.

\bibitem[Li and Cao, 2019]{li2019extended}
Li, Y. and Cao, W. (2019).
\newblock An extended multilayer perceptron model using reduced geometric algebra.
\newblock {\em IEEE Access}, 7:129815--129823.

\bibitem[Maclagan and Sturmfels, 2015]{maclagan2015introduction}
Maclagan, D. and Sturmfels, B. (2015).
\newblock {\em Introduction to {T}ropical {G}eometry}, volume 161.
\newblock American Mathematical Soc.

\bibitem[Makarim and Stevens, 2017]{makarim2017M4GB}
Makarim, R. and Stevens, M. (2017).
\newblock {M4GB}: {A}n efficient {G}röbner-basis algorithm.
\newblock In {\em Proceedings of the International Symposium on Symbolic and Algebraic Computation}, pages 293--300.

\bibitem[Maragos et~al., 2021]{maragos2021tropical}
Maragos, P., Charisopoulos, V., and Theodosis, E. (2021).
\newblock Tropical geometry and machine learning.
\newblock {\em Proceedings of the IEEE}, 109(5):728--755.

\bibitem[Mayr and Meyer, 1982]{mayr1982complexity}
Mayr, E.~W. and Meyer, A.~R. (1982).
\newblock The complexity of the word problems for commutative semigroups and polynomial ideals.
\newblock {\em Advances in Mathematics}, 46(3):305--329.

\bibitem[Mehta et~al., 2022]{mehta2022}
Mehta, D., Chen, T., Tang, T., and Hauenstein, J.~D. (2022).
\newblock The loss surface of deep linear networks viewed through the algebraic geometry lens.
\newblock {\em IEEE Transactions on Pattern Analysis and Machine Intelligence}, 44(9):5664--5680.

\bibitem[Mishkin and Pilanci, 2023]{mishkin2023optimal}
Mishkin, A. and Pilanci, M. (2023).
\newblock Optimal sets and solution paths of {R}e{LU} networks.
\newblock In {\em Proceedings of the 40th International Conference on Machine Learning}, volume 202 of {\em Proceedings of Machine Learning Research}, pages 24888--24924. PMLR.

\bibitem[Mishkin et~al., 2022]{mishkin2022fast}
Mishkin, A., Sahiner, A., and Pilanci, M. (2022).
\newblock Fast convex optimization for two-layer {R}e{LU} networks: Equivalent model classes and cone decompositions.
\newblock In {\em Proceedings of the 39th International Conference on Machine Learning}, volume 162 of {\em Proceedings of Machine Learning Research}, pages 15770--15816. PMLR.

\bibitem[Monagan et~al., 2005]{michael2005maple}
Monagan, M.~B., Geddes, K.~O., Heal, K.~M., Labahn, G., Vorkoetter, S.~M., McCarron, J., and DeMarco, P. (2005).
\newblock {\em Maple~10 Programming Guide}.
\newblock Maplesoft, Waterloo ON, Canada.

\bibitem[Mont\'{u}far et~al., 2014]{Montufar2014}
Mont\'{u}far, G., Pascanu, R., Cho, K., and Bengio, Y. (2014).
\newblock On the number of linear regions of deep neural networks.
\newblock In {\em Advances in Neural Information Processing Systems}, volume~27. Curran Associates, Inc.

\bibitem[Murphy and Van~der Vaart, 2000]{murphy2000profile}
Murphy, S.~A. and Van~der Vaart, A.~W. (2000).
\newblock On profile likelihood.
\newblock {\em Journal of the American Statistical Association}, 95(450):449--465.

\bibitem[Neyshabur et~al., 2018]{Neyshabur2018Towards}
Neyshabur, B., Bhojanapalli, S., McAllester, D., and Srebro, N. (2018).
\newblock A pac-bayesian approach to spectrally-normalized margin bounds for neural networks.
\newblock {\em International Conference on Learning Representations}.

\bibitem[Parada-Mayorga and Ribeiro, 2021]{parada-mayorga2021algebraic}
Parada-Mayorga, A. and Ribeiro, A. (2021).
\newblock Algebraic neural networks: Stability to deformations.
\newblock {\em IEEE Transactions on Signal Processing}, 69:3351--3366.

\bibitem[Peng, 2017]{peng2017multilayerperceptronalgebra}
Peng, Z. (2017).
\newblock Multilayer perceptron algebra.

\bibitem[Pham and Garg, 2024]{pham2024tropicalGNN}
Pham, T.~A. and Garg, V. (2024).
\newblock What do graph neural networks learn? insights from tropical geometry.
\newblock In {\em Advances in Neural Information Processing Systems}, volume~37, pages 10988--11020. Curran Associates, Inc.

\bibitem[Pilanci and Ergen, 2020]{pilanci2020neural}
Pilanci, M. and Ergen, T. (2020).
\newblock Neural networks are convex regularizers: Exact polynomial-time convex optimization formulations for two-layer networks.
\newblock In {\em Proceedings of the 37th International Conference on Machine Learning}, volume 119 of {\em Proceedings of Machine Learning Research}, pages 7695--7705. PMLR.

\bibitem[Reed et~al., 2022]{reed2022generalist}
Reed, S., Zolna, K., Parisotto, E., Colmenarejo, S.~G., Novikov, A., Barth-maron, G., Gim{\'e}nez, M., Sulsky, Y., Kay, J., Springenberg, J.~T., Eccles, T., Bruce, J., Razavi, A., Edwards, A., Heess, N., Chen, Y., Hadsell, R., Vinyals, O., Bordbar, M., and de~Freitas, N. (2022).
\newblock A generalist agent.
\newblock {\em Transactions on Machine Learning Research}.
\newblock Featured Certification, Outstanding Certification.

\bibitem[Ritter and Urcid, 2003]{ritter2003lattice}
Ritter, G.~X. and Urcid, G. (2003).
\newblock Lattice algebra approach to single-neuron computation.
\newblock {\em IEEE Transactions on Neural Networks}, 14(2):282--295.

\bibitem[Rosenblatt, 1958]{rosenblatt1958perceptron}
Rosenblatt, F. (1958).
\newblock {The perceptron: A probabilistic model for information storage and organization in the brain.}
\newblock {\em Psychological Review}, 65(6):386--408.

\bibitem[Shen et~al., 2022]{shen2022optimal}
Shen, Z., Yang, H., and Zhang, S. (2022).
\newblock Optimal approximation rate of relu networks in terms of width and depth.
\newblock {\em Journal de Mathématiques Pures et Appliquées}, 157:101--135.

\bibitem[Vaswani et~al., 2017]{vaswani2017attention}
Vaswani, A., Shazeer, N., Parmar, N., Uszkoreit, J., Jones, L., Gomez, A.~N., Kaiser, L.~u., and Polosukhin, I. (2017).
\newblock Attention is all you need.
\newblock In {\em Advances in Neural Information Processing Systems}, volume~30. Curran Associates, Inc.

\bibitem[Watanabe, 2009]{watanabe2009}
Watanabe, S. (2009).
\newblock {\em Algebraic Geometry and Statistical Learning Theory}.
\newblock Cambridge Monographs on Applied and Computational Mathematics. Cambridge University Press.

\bibitem[Watanabe, 2010]{watanabe2010}
Watanabe, S. (2010).
\newblock Asymptotic equivalence of bayes cross validation and widely applicable information criterion in singular learning theory.
\newblock {\em Journal of Machine Learning Research}, 11(116):3571--3594.

\bibitem[{Wolfram Research{,} Inc.}, 2024]{Mathematica}
{Wolfram Research{,} Inc.} (2024).
\newblock Mathematica, {V}ersion 14.2.
\newblock Champaign, IL, 2024.

\bibitem[Zhang et~al., 2018]{zhang2018tropical}
Zhang, L., Naitzat, G., and Lim, L.-H. (2018).
\newblock Tropical geometry of deep neural networks.
\newblock In {\em Proceedings of the 35th International Conference on Machine Learning}, volume~80 of {\em Proceedings of Machine Learning Research}, pages 5824--5832. PMLR.

\bibitem[Zitkovich et~al., 2023]{zitkovich2023rt2}
Zitkovich, B., Yu, T., Xu, S., Xu, P., Xiao, T., Xia, F., Wu, J., Wohlhart, P., Welker, S., Wahid, A., Vuong, Q., Vanhoucke, V., Tran, H., Soricut, R., Singh, A., Singh, J., Sermanet, P., Sanketi, P.~R., Salazar, G., Ryoo, M.~S., Reymann, K., Rao, K., Pertsch, K., Mordatch, I., Michalewski, H., Lu, Y., Levine, S., Lee, L., Lee, T.-W.~E., Leal, I., Kuang, Y., Kalashnikov, D., Julian, R., Joshi, N.~J., Irpan, A., Ichter, B., Hsu, J., Herzog, A., Hausman, K., Gopalakrishnan, K., Fu, C., Florence, P., Finn, C., Dubey, K.~A., Driess, D., Ding, T., Choromanski, K.~M., Chen, X., Chebotar, Y., Carbajal, J., Brown, N., Brohan, A., Arenas, M.~G., and Han, K. (2023).
\newblock {RT-2}: Vision-language-action models transfer web knowledge to robotic control.
\newblock In {\em Proceedings of The 7th Conference on Robot Learning}, volume 229 of {\em Proceedings of Machine Learning Research}, pages 2165--2183. PMLR.

\end{thebibliography}


\clearpage

\appendix

\renewcommand{\thesection}{S.\arabic{section}} 
\renewcommand{\thesubsection}{S.\arabic{section}.\arabic{subsection}} 
\setcounter{section}{0} 

\section*{Supplementary Material} 
\addcontentsline{toc}{section}{Supplementary Material}

Supplement~\ref{supp:coefficients} shows the rational coefficients used to describe the solutions obtained in our computation. 
Supplement~\ref{supp:illustrative_example} presents an illustrative example of our algorithm, in the simplest case ($L=1$). 
Supplement~\ref{supp:empirical_expression_scaling} presents the actual scale of the algebraic formulae involved in our numerical experiments. Supplement~\ref{supp:basic_concepts} provides additional background on computational algebra.

\section{Coefficients}
\label{supp:coefficients}

In this section, we list the rational coefficients used to describe the one-dimensional solutions obtained in the computations presented in Section~\ref{subsec:results_L=2}.

\begin{align*}
R_1 &= \frac{8061831845311915622677137119327762091177021647160801855468750}{799152119487995315448496053126952456312807787539926712735352143}, \\
R_2 &= \frac{24185495535935746868031411357983286273531064941482405566406250}{799152119487995315448496053126952456312807787539926712735352143}, \\
R_3 &= \frac{16592903810388605869109122181308724918558592156970414314140625}{114164588498285045064070864732421779473258255362846673247907449}, \\
R_4 &= \frac{24185495535935746868031411357983286273531064941482405566406250}{799152119487995315448496053126952456312807787539926712735352143}, \\
R_5 &= \frac{33185807620777211738218244362617449837117184313940828628281250}{114164588498285045064070864732421779473258255362846673247907449}, \\
R_6 &= \frac{3631820373341883747515837259737976349533140846091509424078088125}{913316707986280360512566917859374235786066042902773385983259592}, \\
R_7 &= \frac{8061831845311915622677137119327762091177021647160801855468750}{799152119487995315448496053126952456312807787539926712735352143}, \\
R_8 &= \frac{16592903810388605869109122181308724918558592156970414314140625}{114164588498285045064070864732421779473258255362846673247907449}, \\
R_9 &= \frac{3631820373341883747515837259737976349533140846091509424078088125}{913316707986280360512566917859374235786066042902773385983259592}, \\
R_{10} &= \frac{3986185952593039040079422065453083833713933848669678031131169525}{799152119487995315448496053126952456312807787539926712735352143}, \\
R_{11} &= \frac{8061831845311915622677137119327762091177021647160801855468750}{799152119487995315448496053126952456312807787539926712735352143}, \\
R_{12} &= \frac{24185495535935746868031411357983286273531064941482405566406250}{799152119487995315448496053126952456312807787539926712735352143}, \\
R_{13} &= \frac{16592903810388605869109122181308724918558592156970414314140625}{114164588498285045064070864732421779473258255362846673247907449}, \\
R_{14} &= \frac{24185495535935746868031411357983286273531064941482405566406250}{799152119487995315448496053126952456312807787539926712735352143}, \\
R_{15} &= \frac{33185807620777211738218244362617449837117184313940828628281250}{114164588498285045064070864732421779473258255362846673247907449}, \\
R_{16} &= \frac{3631820373341883747515837259737976349533140846091509424078088125}{913316707986280360512566917859374235786066042902773385983259592}, \\
R_{17} &= \frac{8061831845311915622677137119327762091177021647160801855468750}{799152119487995315448496053126952456312807787539926712735352143}, \\
R_{18} &= \frac{16592903810388605869109122181308724918558592156970414314140625}{114164588498285045064070864732421779473258255362846673247907449}, \\
R_{19} &= \frac{3631820373341883747515837259737976349533140846091509424078088125}{913316707986280360512566917859374235786066042902773385983259592}, \\
R_{20} &= \frac{3986185952593039040079422065453083833713933848669678031131169525}{799152119487995315448496053126952456312807787539926712735352143}, \\
R_{21} &= \frac{91676796916186307}{5836063856703750}, \\
R_{22} &= \frac{91676796916186307}{1945354618901250}, \\
R_{23} &= \frac{10799719744535841949933618669}{26384932237115504306250000}, \\
R_{24} &= \frac{91676796916186307}{1945354618901250}, \\
R_{25} &= \frac{10799719744535841949933618669}{13192466118557752153125000}, \\
R_{26} &= \frac{1170757087686584669238812}{329811652963943803828125}, \\
R_{27} &= \frac{91676796916186307}{5836063856703750}, \\
R_{28} &= \frac{10799719744535841949933618669}{26384932237115504306250000}, \\
R_{29} &= \frac{1170757087686584669238812}{329811652963943803828125}, \\
R_{30} &= \frac{1687032323955370090976492929}{1030661415512324386962890625}.
\end{align*}

\section{Illustrative Example: A Simple Case}
\label{supp:illustrative_example}

In this section, we provide an illustrative example of the proposed Algorithm~\ref{alg:proposal}. To this end, we throughout consider the following fixed dataset consisting of two observations \( \{(x_i, y_i)\}_{i=1}^{n} \subset \mathbb{R}^{d} \times \mathbb{R} \), with \( n = 2 \) and \( d = 1 \):
\[
   (x_1, y_1) = \left( - \frac{17}{100}, - \frac{11}{25} \right), \quad
   (x_2, y_2) = \left( \frac{11}{25}, \frac{19}{20} \right).
\]
To illustrate the basic idea, we consider the case \( L = 1 \), corresponding to a single hidden unit. In this case, the parameters reduce to $\psi=(b_{11},c_1)$ as \( B = (b_{11}) \) and \( c = (c_1) \).

The following exposition is divided into separate sections for clarity, but the content forms a continuous discussion. By the end of this section, we arrive at an example of enumerating the all local minima candidates $\Psi^{\dagger}$.

\subsection{Surrogate Functions}
\label{ex1}

We construct the following indicator matrices in \( \{ \pm 1 \}^{2 \times 1} \):
  \[
     E_1 = 
     \begin{pmatrix}
       +1 & +1
     \end{pmatrix}^{\top},
     ~~~~
     E_2 = 
     \begin{pmatrix}
       +1 & -1
     \end{pmatrix}^{\top},
     ~~~~
     E_3 = 
     \begin{pmatrix}
       -1 & +1
     \end{pmatrix}^{\top},
     ~~~~
     E_4 = 
     \begin{pmatrix}
       -1 & -1
     \end{pmatrix}^{\top}.
  \]
  By using these indicator matrices, we devide the parameter set as \(\Psi = \Psi(E_1) \cup \Psi(E_2) \cup \Psi(E_3) \cup \Psi(E_4)\), where the partitions \(\Psi(E_1), \, \Psi(E_2), \,  \Psi(E_3), \,  \Psi(E_4)\) are defined by 
  \begin{align*}
    \Psi(E_1)
    &=
    \left\{
    \psi \in \Psi 
    \mid 
    (b_{1} x_1 + c_{1}) \times (+1) \ge 0, \,
    (b_{1} x_2 + c_{1}) \times (+1) \ge 0
    \right\}
    \\
    &=
    \left\{
    \psi \in \Psi 
    \mid 
    - \frac{17}{100} b_{1} + c_{1} \ge 0, \, 
    \frac{11}{25} b_{1}  + c_{1} \ge 0
    \right\},
    \\
    \Psi(E_2)
    &=
    \left\{
    \psi \in \Psi 
    \mid 
    (b_{1} x_1 + c_{1}) \times (+1) \ge 0, \,
    (b_{1} x_2 + c_{1}) \times (-1) \ge 0
    \right\}
    \\
    &=
    \left\{
    \psi \in \Psi 
    \mid 
    - \frac{17}{100} b_{1} + c_{1} \ge 0, \,
    \frac{11}{25} b_{1} + c_{1} \leq 0
    \right\},   
    \\
    \Psi(E_3)
    &=
    \left\{
    \psi \in \Psi 
    \mid 
    (b_{1} x_1 + c_{1}) \times (-1) \ge 0, \,
    (b_{1} x_2 + c_{1}) \times (+1) \ge 0
    \right\}
    \\
    &=
    \left\{
    \psi \in \Psi 
    \mid 
    - \frac{17}{100} b_{1} + c_{1}\leq 0, \,
    \frac{11}{25} b_{1} + c_{1}\ge 0
    \right\},
    \\
    \Psi(E_4)
    &=
    \left\{
    \psi \in \Psi 
    \mid 
    (b_{1} x_1 + c_{1}) \times (-1) \ge 0, \,
    (b_{1} x_2 + c_{1}) \times (-1) \ge 0
    \right\}
    \\
    &=
    \left\{
    \psi \in \Psi 
    \mid 
    - \frac{17}{100} b_{1} + c_{1}\leq 0, \,
    \frac{11}{25} b_{1} + c_{1}\leq 0
    \right\},
  \end{align*}
  respectively. In addition, we construct the following linear functions:
  \begin{align*}
    \cov_{1E_1} 
    &= \left(\frac{1}{2} (1 + 1) (b_{1} x_{1} + c_{1})\right)
    = \left(- \frac{17}{100} b_{1} + c_{1}\right),
    &
    \cov_{2E_1} 
    &= \left(\frac{1}{2} (1 + 1) (b_{1} x_{2} + c_{1})\right)
    = \left(- \frac{11}{25} b_{1} + c_{1}\right),
    \\
    \cov_{1E_2} 
    &= \left(\frac{1}{2} (1 + 1) (b_{1} x_{1} + c_{1})\right)
    = \left(- \frac{17}{100} b_{1} + c_{1}\right),
    &
    \cov_{2E_2} 
    &= \left(\frac{1}{2} (-1 + 1) (b_{1} x_{2} + c_{1})\right)
    = \left(0\right),
    \\
    \cov_{1E_3} 
    &= \left(\frac{1}{2} (- 1 + 1) (b_{1} x_{1} + c_{1})\right)
    = \left(0\right),
    &
    \cov_{2E_3} 
    &= \left(\frac{1}{2} (1 + 1) (b_{1} x_{2} + c_{1})\right)
    = \left(- \frac{11}{25} b_{1} + c_{1}\right),
    \\
    \cov_{1E_4} 
    &= \left(\frac{1}{2} (- 1 + 1) (b_{1} x_{1} + c_{1})\right)
    = \left(0\right),
    &
    \cov_{2E_4} 
    &= \left(\frac{1}{2} (-1 + 1) (b_{1} x_{2} + c_{1})\right)
    = \left(0\right).
  \end{align*}
  Let 
  \[
  \COV_{E_1} = (\cov_{1E_1}^{\top}, \cov_{2E_1}^{\top})^{\top}, \, \COV_{E_2} = (\cov_{1E_2}^{\top}, \cov_{2E_2}^{\top})^{\top} , \, \COV_{E_3} = (\cov_{1E_3}^{\top}, \cov_{2E_3}^{\top})^{\top}, \, \COV_{E_4} = (\cov_{1E_4}^{\top}, \cov_{2E_4}^{\top})^{\top} \in \mathbb{R}^{2 \times 1}. 
  \]
  Using these matrices, we construct the algebraic surrogates in the case \(\lambda = 0.1\). 
  These surrogates for the partitions \(\Psi(E_1), \Psi(E_2), \Psi(E_3), \Psi(E_4)\) are:
  {\small\begin{align*}
    \ell_{0.1,E_1}(\psi) 
    &=
    \frac{1}{10} \|\psi\|^2
    -
    \ip{\widetilde{y}}{\COV_{E1}(\COV_{E_1}^{\top} \COV_{E_1} + \lambda I_L)^{-1} \COV_{E_1}^{\top} \widetilde{y}} 
    \\
    &=
    \frac{\frac{89}{256000000000} b_{11}^4 + \frac{27}{32000000000} b_{11}^3 c_1 + \frac{889}{256000000000} b_{11}^2 c_1^2 + \frac{27}{32000000000} b_{11} c_1^3 + \frac{1}{320000000} c_1^4}{D_1}
    \\
    &\hspace*{4.3cm} +\frac{- \frac{67893441}{25600000000000000} b_{11}^2 + \frac{1}{6400000000} c_1^2}{D_1},
    \\
    \ell_{0.1,E_2}(\psi) 
    &=
    \frac{1}{10} \|\psi\|^2
    -
    \ip{\widetilde{y}}{\COV_{E2}(\COV_{E_2}^{\top} \COV_{E_2} + \lambda I_L)^{-1} \COV_{E_2}^{\top} \widetilde{y}} 
    \\
    &=
    \frac{\frac{289}{100000} b_{11}^4 - \frac{17}{500} b_{11}^3 c_1 + \frac{10289}{100000} b_{11}^2 c_1^2 - \frac{17}{500} b_{11} c_1^3 + \frac{1}{10} c_1^4 - \frac{1583769}{400000000} b_{11}^2 + \frac{328457}{2000000} b_{11} c_1 - \frac{18921}{40000} c_1^2}{D_2},
    \\
    \ell_{0.1,E_3}(\psi) 
    &=
    \frac{1}{10} \|\psi\|^2
    -
    \ip{\widetilde{y}}{\COV_{E1}(\COV_{E_3}^{\top} \COV_{E_3} + \lambda I_L)^{-1} \COV_{E_3}^{\top} \widetilde{y}} 
    \\
    &=
    \frac{\frac{121}{6250} b_{11}^4 + \frac{11}{125} b_{11}^3 c_1 + \frac{373}{3125} b_{11}^2 c_1^2 + \frac{11}{125} b_{11} c_1^3 + \frac{1}{10} c_1^4 - \frac{2087841}{25000000} b_{11}^2 - \frac{212531}{500000} b_{11} c_1 - \frac{18921}{40000} c_1^2}{D_3},
    \\
    \ell_{0.1,E_4}(\psi) 
    &=
    \frac{1}{10} \|\psi\|^2
    -
    \ip{\widetilde{y}}{\COV_{E_4}(\COV_{E_4}^{\top} \COV_{E_4} + \lambda I_L)^{-1} \COV_{E_4}^{\top} \widetilde{y}} 
    \\
    &=
    \frac{1}{10} b_{11}^2 + \frac{1}{10} c_1^2,
  \end{align*}}
  where
  \begin{align*}
    D_1 &= \frac{89}{25600000000} b_{11}^2 + \frac{27}{3200000000} b_{11} c_1 + \frac{1}{32000000} c_1^2 + \frac{1}{640000000},
    \\
    D_2 &= \frac{289}{10000} b_{11}^2 - \frac{17}{50} b_{11} c_1 + c_1^2 + \frac{1}{10},
    \\
    D_3 &= \frac{121}{625} b_{11}^2 + \frac{22}{25} b_{11} c_1 + c_1^2 + \frac{1}{10}.
  \end{align*}

\subsection{Candidates Inside the Partition Interior}
\label{ex2}

To enumerate local minima candidates, we define the interior of each partition \(\Psi(E_i)\):
  \begin{align*}
     \Phi(E_i) = \Psi(E_i) \cap \left\{ \psi \in \Psi \mid - \frac{17}{100} b_{1} + c_{1} \neq 0, \, \frac{11}{25} b_{1} + c_{1} \neq 0 \right\}.
  \end{align*}
  We then construct the gradient of the algebraic surrogate \(\ell_{0.1,E_i}\) for each partition \(\Psi(E_i)\).
  {\scriptsize\begin{align*}
     \frac{\partial \ell_{0.1,E_1}}{\partial \psi} &= 
     \left\{
     \begin{array}{l}
     \frac{\frac{7921}{3276800000000000000000} b_{11}^5 + \frac{2403}{204800000000000000000} b_{11}^4 c_1 + \frac{1477}{25600000000000000000} b_{11}^3 c_1^2}{D_1^2},
     \\
    \hspace*{1cm} + \: \frac{\frac{27}{256000000000000000} b_{11}^2 c_1^3 + \frac{1}{5120000000000000} b_{11} c_1^4 + \frac{89}{40960000000000000000} b_{11}^3}{D_1^2}
     \\
    \hspace*{2cm} - \: \frac{\frac{1509122907}{81920000000000000000000000} b_{11}^2 c_1 + \frac{63893441}{409600000000000000000000} b_{11} c_1^2 + \frac{67893441}{8192000000000000000000000} b_{11}}{D_1^2},
    \\
     \frac{\frac{7921}{3276800000000000000000} b_{11}^4 c_1 + \frac{2403}{204800000000000000000} b_{11}^3 c_1^2 + \frac{1477}{25600000000000000000} b_{11}^2 c_1^3}{D_1^2}
    \\
     \hspace*{1cm} + \: \frac{\frac{27}{256000000000000000} b_{11} c_1^4 + \frac{1}{5120000000000000} c_1^5 + \frac{1941122907}{81920000000000000000000000} b_{11}^3}{D_1^2}
    \\
     \hspace*{2cm} + \: \frac{\frac{72783441}{409600000000000000000000} b_{11}^2 c_1 + \frac{27}{5120000000000000000} b_{11} c_1^2 + \frac{1}{51200000000000000} c_1^3 + \frac{1}{2048000000000000000} c_1}{D_1^2}
     \end{array}
     \right\}
     \\
     &=:
     \left\{\frac{f_{11}}{D_1^2}, \frac{f_{21}}{D_1^2} \right\},
     \\
     \frac{\partial \ell_{0.1,E_2}}{\partial \psi} &= \left\{
     \begin{array}{l}
     \frac{\frac{83521}{500000000} b_{11}^5 - \frac{4913}{1250000} b_{11}^4 c_1 + \frac{867}{25000} b_{11}^3 c_1^2 - \frac{17}{125} b_{11}^2 c_1^3 + \frac{1}{5} b_{11} c_1^4 + \frac{289}{250000} b_{11}^3 - \frac{17}{1250} b_{11}^2 c_1 + \frac{1}{25} b_{11} c_1^2 - \frac{1583769}{2000000000} b_{11} + \frac{328457}{20000000} c_1}{D_2^2}
     \\
     \frac{\frac{83521}{500000000} b_{11}^4 c_1 - \frac{4913}{1250000} b_{11}^3 c_1^2 + \frac{867}{25000} b_{11}^2 c_1^3 - \frac{17}{125} b_{11} c_1^4 + \frac{1}{5} c_1^5 + \frac{289}{250000} b_{11}^2 c_1 - \frac{17}{1250} b_{11} c_1^2 + \frac{1}{25} c_1^3 + \frac{328457}{20000000} b_{11} - \frac{18921}{200000} c_1}{D_2^2}
     \end{array}
     \right\}
    \\
     &=:
     \left\{\frac{f_{12}}{D_2^2}, \frac{f_{22}}{D_2^2} \right\},
     \\
     \frac{\partial \ell_{0.1,E_3}}{\partial \psi} &= \left\{
     \begin{array}{l}
     \frac{\frac{14641}{1953125} b_{11}^5 + \frac{5324}{78125} b_{11}^4 c_1 + \frac{726}{3125} b_{11}^3 c_1^2 + \frac{44}{125} b_{11}^2 c_1^3 + \frac{1}{5} b_{11} c_1^4 + \frac{121}{15625} b_{11}^3 + \frac{22}{625} b_{11}^2 c_1 + \frac{1}{25} b_{11} c_1^2 - \frac{2087841}{125000000} b_{11} - \frac{212531}{5000000} c_1}{D_3^2},
     \\
     \frac{\frac{14641}{1953125} b_{11}^4 c_1 + \frac{5324}{78125} b_{11}^3 c_1^2 + \frac{726}{3125} b_{11}^2 c_1^3 + \frac{44}{125} b_{11} c_1^4 + \frac{1}{5} c_1^5 + \frac{121}{15625} b_{11}^2 c_1 + \frac{22}{625} b_{11} c_1^2 + \frac{1}{25} c_1^3 - \frac{212531}{5000000} b_{11} - \frac{18921}{200000} c_1}{D_3^2}
     \end{array}
     \right\}
    \\
     &=:
     \left\{\frac{f_{13}}{D_3^2}, \frac{f_{23}}{D_3^2} \right\},
     \\
     \frac{\partial \ell_{0.1,E_4}}{\partial \psi} &= \left\{\frac{1}{5} b_{11}, \frac{1}{5} c_1\right\} =: \{ f_{14}, f_{24}\}.
  \end{align*}}
For notational consistency in the presentation of algebraic equation systems, we set \(D_4 = 1\). 
    
We now enumerate the stationary points of the first algebraic surrogate, \( \ell_{0.1, E_1} \).
To this end, we consider the estimating equation \( \partial \ell_{0.1, E_1} / \partial \psi = 0 \), and characterize stationary points as the solutions to the following system:
  \begin{align}\label{eq:ex:or1}
    \begin{cases}
       0 = f_{1i},
       \\
       0 = f_{2i},
       \\
       0 \neq D_i,
       \\
       0 \neq - \frac{17}{100} b_{1} + c_{1},
       \\
       0 \neq \frac{11}{25} b_{1} + c_{1}.
    \end{cases}
  \end{align}
Therein, the third condition \( 0 \ne D_i \) ensures that the denominator of the derivative \( \partial \ell_{0.1,E_1} / \partial \psi \) is non-zero, while the fourth and fifth conditions guarantee that the point \( \psi \) does not lie on the boundary. 
If we identify all parameters \( b_{11}, c_1 \) that satisfy the conditions within the domain \( \Psi(E_i) \), then we obtain all stationary points of the algebraic surrogate \( \ell_{0.1, E_i} \) that lie in the interior \( \Phi(E_i) \).
 
Since we have the equivalent relation
  \[
    \begin{cases}
       0 \neq D_i,
       \\
       0 \neq - \frac{17}{100} b_{1} + c_{1},
       \\
       0 \neq \frac{11}{25} b_{1} + c_{1}
    \end{cases}
    \quad
    \Longleftrightarrow
    \quad\quad
    0 \neq D_i \times \left(- \frac{17}{100} b_{1} + c_{1}\right) \left(\frac{11}{25} b_{1} + c_{1}\right),
  \]
  we characterize the following difference set:
  \[
    V_i = \mathbb{V}(f_{1i}, f_{2i}) \setminus \mathbb{V}\left(D_i \times \left(- \frac{17}{100} b_{1} + c_{1}\right) \times \left(\frac{11}{25} b_{1} + c_{1}\right)\right).
  \]

  Following the remaining technical calculation presented in Supplement~\ref{subsec:remaining_technical_computations}, we obtain
  \begin{align}
  V_i \cap \Psi(E_i) = \emptyset
    \label{eq:solutions_interior}
  \end{align}
  for each \(i\).

  \subsection{Candidates on the Partition Boundary}
\label{ex3}

  This section describes the local minima candidates on the partition boundaries. 
  Firstly, define boundaries of each parameter space \(\Psi(E_i)\):
  \begin{align*}
     \Gamma_1(E_i) &= \Psi(E_i) \cap \left\{ \psi \in \Psi \mid - \frac{17}{100} b_{1} + c_{1} = 0 \right\},
     \\
     \Gamma_2(E_i) &= \Psi(E_i) \cap \left\{ \psi \in \Psi \mid \frac{11}{25} b_{1} + c_{1} = 0 \right\}.
  \end{align*}
  We can find local minima candidates, by leveraging the following Lagrange functions:
  \begin{align*}
    L_{1, E_i} &= \ell_{0.1,E_i} + \beta \left( - \frac{17}{100} b_{1} + c_{1} \right),
    \\
    L_{2, E_i} &= \ell_{0.1,E_i} + \beta \left( \frac{11}{25} b_{1} + c_{1} \right),
  \end{align*}
  where \(\beta\) is a Lagrange multiplier. We denote the denominator and numerator of a rational expression by \( \den(\cdot) \) and \( \num(\cdot) \), respectively. Then, we define: 
  \begin{align*}
    W_{1i} = \mathbb{V}\left(\mathrm{num}\left(\frac{\partial L_{1,E_i}}{\partial b_{11}}\right), \mathrm{num}\left(\frac{\partial L_{1,E_i}}{\partial c_{1}}\right), \mathrm{num}\left(\frac{\partial L_{1,E_i}}{\partial \beta}\right) \right) \setminus \mathbb{V}\left(\mathrm{den}\left(\frac{\partial L_{1,E_i}}{\partial b_{11}}\right) \times \mathrm{den}\left(\frac{\partial L_{1,E_i}}{\partial c_{1}}\right)\right),
    \\
    W_{2i} = \mathbb{V}\left(\mathrm{num}\left(\frac{\partial L_{2,E_i}}{\partial b_{11}}\right), \mathrm{num}\left(\frac{\partial L_{2,E_i}}{\partial c_{1}}\right), \mathrm{num}\left(\frac{\partial L_{2,E_i}}{\partial \beta}\right) \right) \setminus \mathbb{V}\left(\mathrm{den}\left(\frac{\partial L_{2,E_i}}{\partial b_{11}}\right) \times \mathrm{den}\left(\frac{\partial L_{2,E_i}}{\partial c_{1}}\right)\right),
  \end{align*}
  as in \eqref{eq:difference-2}. 
Following the remaining technical calculation presented in Supplement~\ref{subsec:remaining_technical_computations}, we obtain the following algebraic solutions consequently:
\begin{align}
    \bigcup_{i=1}^4 \left(W_{1i} \cup W_{2i}\right) \cap \Psi(E_i)
    =
    \left\{
        (0, 0),
        (0.92, 0.16),
        (-0.88, 0.39)
    \right\}.
    \label{eq:solutions_boundary}
\end{align}

\subsection{Remaining Technical Computations}
\label{subsec:remaining_technical_computations}

This final section completes our analysis by presenting the remaining technical computations.

We define the following ideals:
  \begin{align*}
    \mathcal{I}_i &= \ideal{f_{1i},f_{2i}} \\
    \mathcal I_{1i} &= \ideal{\mathrm{num}\left(\frac{\partial L_{1,E_i}}{\partial b_{11}}\right), \mathrm{num}\left(\frac{\partial L_{1,E_i}}{\partial c_{1}}\right), \frac{\partial L_{1,E_i}}{\partial \beta}},
    \\
    \mathcal I_{2i} &= \ideal{\mathrm{num}\left(\frac{\partial L_{2,E_i}}{\partial b_{11}}\right), \mathrm{num}\left(\frac{\partial L_{2,E_i}}{\partial c_{1}}\right), \frac{\partial L_{2,E_i}}{\partial \beta}}, \\
      \mathcal{K}_i &= \ideal{D_i \times \left(- \frac{17}{100} b_{1} + c_{1}\right) \times \left(\frac{11}{25} b_{1} + c_{1}\right)},
      \\
      \mathcal{K}_{1,i} &= \ideal{\mathrm{den}\left(\frac{\partial L_{1,E_i}}{\partial b_{11}}\right) \times \mathrm{den}\left(\frac{\partial L_{1,E_i}}{\partial c_{1}}\right)},
      \\
      \mathcal{K}_{2,i} &= \ideal{\mathrm{den}\left(\frac{\partial L_{2,E_i}}{\partial b_{11}}\right) \times \mathrm{den}\left(\frac{\partial L_{2,E_i}}{\partial c_{1}}\right)}.
   \end{align*}

The generators of the saturation and elimination ideals are derived by computing the appropriate Gr{\"o}bner bases. In the current setting, we obtain:
   \begin{align*}
       \mathcal{I}_1:\mathcal{K}_1 &= \ideal{h_1, \, h_2}
       \\
       \mathcal{I}_2:\mathcal{K}_2 &= \ideal{b_{11} + \frac{17}{100} c_{1}, c_{1}^{4} + \frac{20000000}{105863521} c_{1}^{2} - \frac{4869844225000000}{11207085078517441}},
       \\
       \mathcal{I}_3:\mathcal{K}_3 &= \ideal{b_{11} - \frac{11}{25} c_{1}, c_{1}^{4} + \frac{78125}{556516} c_{1}^{2} - \frac{2766301953125}{9910721864192}},
       \\
       \mathcal{I}_4:\mathcal{K}_4 &= \ideal{1},
       \\
       (\mathcal{I}_{11}:\mathcal{K}_{11})_1 &= \ideal{b_{11} - \frac{100}{17} c_{1}, c_{1}^{5} + \frac{289}{18605} c_{1}^{3} - \frac{5660873058161}{5698394321960000} c_{1}},
       \\
       (\mathcal{I}_{21}:\mathcal{K}_{21})_1 &= \ideal{b_{11} + \frac{25}{11} c_{1}, c_{1}^{5} + \frac{1936}{18605} c_{1}^{3} - \frac{982689879281}{25822493465000} c_{1}},
       \\
       (\mathcal{I}_{12}:\mathcal{K}_{12})_1 &= \ideal{b_{11}, c_{1}},
       \\
       (\mathcal{I}_{22}:\mathcal{K}_{22})_1 &= \ideal{b_{11} + \frac{25}{11} c_{1}, c_{1}^{5} + \frac{1936}{18605} c_{1}^{3} - \frac{982689879281}{25822493465000} c_{1}},
       \\
       (\mathcal{I}_{13}:\mathcal{K}_{13})_1 &= \ideal{b_{11} - \frac{100}{17} c_{1}, c_{1}^{5} + \frac{289}{18605} c_{1}^{3} - \frac{5660873058161}{5698394321960000} c_{1}},
       \\
       (\mathcal{I}_{23}:\mathcal{K}_{23})_1 &= \ideal{b_{11}, c_{1}},
       \\
       (\mathcal{I}_{14}:\mathcal{K}_{14})_1 &= \ideal{b_{11}, c_{1}},
       \\
       (\mathcal{I}_{24}:\mathcal{K}_{24}))_1 &= \ideal{b_{11}, c_{1}},
   \end{align*}
   where 
   \begin{align*}
      h_1 &=
       b_{11} + \; \frac{345876476985085329715723948195404688220312500000000}{19031747656360561416272393983476382389744845661183} c_{1}^{7} \\ 
       &\hspace{5em} - \; \frac{35559747196837883065628777274588348326346882250000000}{1541571560165205474718063912661586973569332498555823} c_{1}^{5} 
       \\
       &\hspace{5em} + \; \frac{446063344696623004585906133541090341242104426539087500}{124867296373381643452163176925588544859115932383021663} c_{1}^{3}        \\
       &\hspace{5em} + \; \frac{1290255906217763301799679314377701823418521305030950}{171285728907245052746451545851287441507703610950647} c_{1}, \\
       h_2 &=
       c_{1}^{8} - \frac{77386512682238381}{43770478925278125} c_{1}^{6} + \frac{317912756385193489269716273}{329811652963943803828125000} c_{1}^{4} \\
       &\hspace{5em} + \; \frac{3775220455209057594259}{144773290738631025000000} c_{1}^{2} - \frac{25349446426759004477574801}{16085921193181225000000000000}.
   \end{align*}

\subsubsection*{(1) Interior}
Firstly, we compute the stationary points in the interior region. 
By the fact that any Gr{\"o}bner basis generates its ideal, Equation~\eqref{eq:ex:or1} is equivalent to the union of systems defined by the Gr{\"o}bner basis generators of \(\mathcal{I}_1:\mathcal{K}_1^{\infty}\). Therefore, we obtain the conditions
  \begin{align}\label{eq:ex:oreq1-1}
    i = 1 ~~ &: ~~
    \begin{cases}
       0 = h_1
       \\
       0 = h_2
       \\
       0 \neq D_1,
       \\
       0 \neq - \frac{17}{100} b_{1} + c_{1},
       \\
       0 \neq \frac{11}{25} b_{1} + c_{1},
    \end{cases}
    \end{align}
Similarly for $i=2,3,4$, 
leveraging the Gr{\"o}bner basis generators of \(\mathcal{I}_i:\mathcal{K}_i^{\infty}\), we obtain the conditions
    \begin{align}
    \label{eq:ex:oreq1-2}
    i = 2 ~~ &: ~~
    \begin{cases}
      0 = b_{11} + \frac{17}{100} c_{1},
      \\
      0 = c_{1}^{4} + \frac{20000000}{105863521} c_{1}^{2} - \frac{4869844225000000}{11207085078517441},
       \\
       0 \neq D_2,
       \\
       0 \neq - \frac{17}{100} b_{1} + c_{1},
       \\
       0 \neq \frac{11}{25} b_{1} + c_{1},
    \end{cases}
    \\
    \label{eq:ex:oreq1-3}
    i = 3 ~~ &: ~~
    \begin{cases}
      0 = b_{11} - \frac{11}{25} c_{1}, 
      \\
      0 = c_{1}^{4} + \frac{78125}{556516} c_{1}^{2} - \frac{2766301953125}{9910721864192},
       \\
       0 \neq D_2,
       \\
       0 \neq - \frac{17}{100} b_{1} + c_{1},
       \\
       0 \neq \frac{11}{25} b_{1} + c_{1},
    \end{cases}
    \\
    \label{eq:ex:oreq1-4}
    i = 4 ~~ &: ~~
    \begin{cases}
      0 = 1, 
       \\
       0 \neq D_2,
       \\
       0 \neq - \frac{17}{100} b_{1} + c_{1},
       \\
       0 \neq \frac{11}{25} b_{1} + c_{1}.
    \end{cases}
  \end{align}

Solving the above conditions, the solution sets of Equations~\eqref{eq:ex:oreq1-1}--\eqref{eq:ex:oreq1-4} are summarized below: 
  \begin{align*}
    \eqref{eq:ex:oreq1-1} ~~ &: ~~ \emptyset,
    \\
    \eqref{eq:ex:oreq1-2} ~~ &: ~~ \emptyset,
    \\
    \eqref{eq:ex:oreq1-3} ~~ &: ~~ \emptyset,
    \\
    \eqref{eq:ex:oreq1-4} ~~ &: ~~ \emptyset.
  \end{align*}
Therefore, there is no stationary point inside the interior; \eqref{eq:solutions_interior} is proved.

\subsubsection*{(2) Boundary}

Similarly, for the boundary case, we derive the necessary conditions that any local minimum must satisfy:
  \begin{align}\label{eq:ex:oreq2-1}
    i = 1 ~~ &: ~~
    \begin{cases}
       0 = b_{11} - \frac{100}{17} c_{1},
       \\
       0 = c_{1}^{5} + \frac{289}{18605} c_{1}^{3} - \frac{5660873058161}{5698394321960000} c_{1},
       \\
       0 \neq \mathrm{den}\left( \frac{\partial L_{1,E_1}}{\partial b_{11}} \right),
       \\
       0 \neq \mathrm{den}\left( \frac{\partial L_{1,E_1}}{\partial c_{1}} \right),
    \end{cases}
    \mbox{ or } ~~~~~~
    \begin{cases}
       0 = b_{11} + \frac{25}{11} c_{1},
       \\
       0 = c_{1}^{5} + \frac{1936}{18605} c_{1}^{3} - \frac{982689879281}{25822493465000} c_{1},
       \\
       0 \neq \mathrm{den}\left( \frac{\partial L_{2,E_1}}{\partial b_{11}} \right),
       \\
       0 \neq \mathrm{den}\left( \frac{\partial L_{2,E_1}}{\partial c_{1}} \right),
    \end{cases}
    \\
    \label{eq:ex:oreq2-2}
    i = 2 ~~ &: ~~
    \begin{cases}
       0 = b_{11},
       \\
       0 = c_{1},
       \\
       0 \neq \mathrm{den}\left( \frac{\partial L_{1,E_2}}{\partial b_{11}} \right),
       \\
       0 \neq \mathrm{den}\left( \frac{\partial L_{1,E_2}}{\partial c_{1}} \right),
    \end{cases}
    \hspace*{3.1cm}
    \mbox{ or } ~~~~~~
    \begin{cases}
       0 = b_{11} + \frac{25}{11} c_{1},
       \\
       0 = c_{1}^{5} + \frac{1936}{18605} c_{1}^{3} - \frac{982689879281}{25822493465000} c_{1},
       \\
       0 \neq \mathrm{den}\left( \frac{\partial L_{2,E_2}}{\partial b_{11}} \right),
       \\
       0 \neq \mathrm{den}\left( \frac{\partial L_{2,E_2}}{\partial c_{1}} \right),
    \end{cases}
    \\
    \label{eq:ex:oreq2-3}
    i = 3 ~~ &: ~~
    \begin{cases}
       0 = b_{11} - \frac{100}{17} c_{1},
       \\
       0 = c_{1}^{5} + \frac{289}{18605} c_{1}^{3} - \frac{5660873058161}{5698394321960000} c_{1},
       \\
       0 \neq \mathrm{den}\left( \frac{\partial L_{1,E_3}}{\partial b_{11}} \right),
       \\
       0 \neq \mathrm{den}\left( \frac{\partial L_{1,E_3}}{\partial c_{1}} \right),
    \end{cases}
    \mbox{ or } ~~~~~~
    \begin{cases}
       0 = b_{11},
       \\
       0 = c_{1},
       \\
       0 \neq \mathrm{den}\left( \frac{\partial L_{2,E_3}}{\partial b_{11}} \right),
       \\
       0 \neq \mathrm{den}\left( \frac{\partial L_{2,E_3}}{\partial c_{1}} \right),
    \end{cases}
    \\
    \label{eq:ex:oreq2-4}
    i = 4 ~~ &: ~~
    \begin{cases}
       0 = b_{11},
       \\
       0 = c_{1},
       \\
       0 \neq \mathrm{den}\left( \frac{\partial L_{1,E_4}}{\partial b_{11}} \right),
       \\
       0 \neq \mathrm{den}\left( \frac{\partial L_{1,E_4}}{\partial c_{1}} \right),
    \end{cases}
    \hspace*{3.1cm}
    \mbox{ or } ~~~~~~
    \begin{cases}
       0 = b_{11},
       \\
       0 = c_{1},
       \\
       0 \neq \mathrm{den}\left( \frac{\partial L_{2,E_4}}{\partial b_{11}} \right),
       \\
       0 \neq \mathrm{den}\left( \frac{\partial L_{2,E_4}}{\partial c_{1}} \right).
    \end{cases}
  \end{align}

By solving the above conditions, we obtain the solutions: 
  \begin{align*}
    \eqref{eq:ex:oreq2-1} ~~ &: ~~ 
    \left\{\
    \begin{array}{ll}
       \left(\frac{100}{17} \sqrt{-\frac{289}{37210} + \frac{40171}{12200 \sqrt{10289}}}\right., & \left. \sqrt{-\frac{289}{37210} + \frac{40171}{12200 \sqrt{10289}}}\right),
       \\
       \left(- \frac{25}{11} \sqrt{-\frac{968}{18605} + \frac{16819}{3050 \sqrt{746}}}\right., & \left.  \sqrt{-\frac{968}{18605} + \frac{16819}{3050 \sqrt{746}}}\right)
    \end{array}
    \right\},
    \\
    \eqref{eq:ex:oreq2-2} ~~ &: ~~ 
    \left\{\
    \begin{array}{ll}
       \left(- \frac{25}{11} \sqrt{-\frac{968}{18605} + \frac{16819}{3050 \sqrt{746}}}\right., & \left.  \sqrt{-\frac{968}{18605} + \frac{16819}{3050 \sqrt{746}}}\right)
    \end{array}
    \right\},
    \\
    \eqref{eq:ex:oreq2-3} ~~ &: ~~ 
    \left\{\
    \begin{array}{ll}
       \left(
           \frac{100}{17} \sqrt{-\frac{289}{37210} + \frac{40171}{12200 \sqrt{10289}}}\right., & \left.  \sqrt{-\frac{289}{37210} + \frac{40171}{12200 \sqrt{10289}}}
       \right)
    \end{array}
    \right\},
    \\
    \eqref{eq:ex:oreq2-4} ~~ &: ~~ 
    \left\{\
    \begin{array}{ll}
       \left(
           0 \right., & \left.  0
       \right)
    \end{array}
    \right\}.
  \end{align*}
By intersecting the resulting solution sets with the partitions \(\Psi(E_1), \Psi(E_2), \Psi(E_3), \Psi(E_4)\), we obtain the candidates for local minima: \eqref{eq:solutions_boundary} is proved.

\section{Empirical Analysis of Symbolic Expression Scaling}

\label{supp:empirical_expression_scaling}

Let \( n = 5 \), \( d = 1 \), and consider the dataset given by
\begin{align*}
x_1 &= \frac{1}{5}, \quad x_2 = \frac{2}{5}, \quad x_3 = \frac{3}{5}, \quad x_4 = \frac{4}{5}, \quad x_5 = 1, \\
y_1 &= 1, \quad y_2 = 0, \quad y_3 = 1, \quad y_4 = 1, \quad y_5 = 0, \quad \lambda = \frac{1}{10}.
\end{align*}

For illustrative purposes, we fix a single instance of indicator matrices \( E^{(L)} \in \{\pm 1\}^{n \times L} \) for each number of hidden units \( L = 1, 2, 3 \), where each entry is chosen from \( \{ \pm 1 \} \). Specifically, we consider the following example matrices:
\begin{align}
E^{(1)} &=
\begin{pmatrix}
 1 \\
 1 \\
-1 \\
-1 \\
-1 \\
\end{pmatrix}, \qquad
E^{(2)} =
\begin{pmatrix}
 1 &  1 \\
-1 & -1 \\
-1 &  1 \\
 1 & -1 \\
-1 &  1 \\
\end{pmatrix}, \qquad
E^{(3)} =
\begin{pmatrix}
 1 &  1 & -1 \\
-1 & -1 &  1 \\
 1 & -1 & -1 \\
 1 &  1 & -1 \\
-1 &  1 &  1 \\
\end{pmatrix}.
\label{eq:instance_of_E}
\end{align}

It should be emphasized that each \( E^{(L)} \) shown above is only one realization from the total space of possible indicator matrices \( \{ \pm 1 \}^{n \times L} \), whose cardinality is \( 2^{nL} \). In practice, our algorithm systematically generates all such \( 2^{nL} \) candidates and solves the corresponding algebraic equations simultaneously for each configuration.

In this section, we focus on the univariate case \( d = 1 \), where the weight matrix \( B = (b_{ij}) \in \mathbb{R}^{L \times d} \) has only one column. Therefore, we abbreviate the entry as a single-indexed real-value, i.e., \( B = (b_1, b_2, \ldots, b_L) \in \mathbb{R}^L \). 
To illustrate the actual size of the resulting equations, we expand below the numerator of the derivative of the R$^3$-MSE with respect to the parameter $b_1$. Derivatives with respect to other parameters such as $b_2,b_3,\ldots,b_L,c_1,c_2,\ldots,c_L$, as well as the corresponding denominators, also appear in the full system, but they are omitted here to save space, as they follow a similar structure and scale with a comparable number of terms.

\subsection{Varying Number of Hidden Units}
\label{supp:varying_width}

In this section, we fix the sample size at $n=5$ and observe how the specific instances of the algebraic equations we handle change as the number of hidden units $L$ varies with $L=1,2,3$. 

As $L$ increases, the number of parameters ($b_1,b_2,\ldots,b_L,c_1,c_2,\ldots,c_L$) involved in the model also increases, thereby enlarging the set of variables that appear in the algebraic equations. Consequently, the number of terms in these equations grows rapidly. Roughly speaking, the number of terms depends on the combinations of parameters appearing in polynomial expressions, and thus increases at a rate far beyond linear as $L$ becomes larger.

\subsubsection*{Case I: $L=1, n=5$}

$\num\bigg( \frac{\partial \ell_{\lambda,E^{(1)}}(\psi)}{\partial b_{1}} \bigg)=$
\begin{quote} \raggedright
$\input{long_equations/L1}$. 
\end{quote}

\subsubsection*{Case II: $L=2, n=5$}

$\num\bigg( \frac{\partial \ell_{\lambda,E^{(2)}}(\psi)}{\partial b_{1}} \bigg)=$
\begin{quote} \raggedright
$\input{long_equations/L2}$. 
\end{quote}

\subsubsection*{Case III: $L=3, n=5$}

$\num\bigg( \frac{\partial \ell_{\lambda,E^{(3)}}(\psi)}{\partial b_{1}} \bigg)=$
\begin{quote} \raggedright
$\input{long_equations/L3}$. 
\end{quote}

\subsection{Varying Sample Size}
\label{supp:varying_sample_size}

In this section, we investigate how the specific instances of the algebraic equations change when the sample size \( n \) is reduced, while keeping other conditions fixed, based on the setting \( L = 3, n = 5 \) described in Supplement~\ref{supp:varying_width} Case III. In particular, for \( n = 4 \), we use the first four data points \(\{(x_1,y_1), (x_2,y_2), (x_3,y_3), (x_4,y_4)\}\) from the dataset used in Supplement~\ref{supp:varying_width}, and define the new indicator matrix as the top \(4 \times 3\) submatrix of \(E^{(3)} \in \{\pm 1\}^{5 \times 3}\) given in \eqref{eq:instance_of_E}. The same procedure is applied for \( n = 3 \) and \( n = 2 \).

The R\(^3\)-MSE can also be interpreted as a prediction problem where the response \( y_i \) is approximated by \(\omega_{i\ell,E}(\psi) = \frac{e_{i\ell}+1}{2}(b_{\ell}x_i + c_{\ell})\). Among these, when \( e_{i\ell} = -1 \), we have \(\omega_{i\ell,E}(\psi) = 0\), meaning that only the terms corresponding to \( e_{i\ell} = +1 \) in \( E=(e_{i\ell}) \in \{\pm 1\}^{n \times L} \) contribute to the overall system of algebraic equations. As \( n \) decreases, the total number of positive elements \( e_{i\ell} = +1 \) in the \( n \times L \) matrix \( E \) tends to decrease as well. Consequently, as the sample size decreases, fewer parameters are involved in the equations, resulting in a simpler overall algebraic structure.
Also note that the number of possible indicator matrices $E$ is $2^{nL}$, so the overall computational complexity decreases exponentially as $n$ becomes smaller.

\subsubsection*{Case IV: $L=3, n=4$}

$\num\bigg( \frac{\partial \ell_{\lambda,E^{(3)}}(\psi)}{\partial b_{1}} \bigg)=$
\begin{quote} \raggedright
$\input{long_equations/L3_n4}$. 
\end{quote}

\subsubsection*{Case V: $L=3, n=3$}

$\num\bigg( \frac{\partial \ell_{\lambda,E^{(3)}}(\psi)}{\partial b_{1}} \bigg)=$
\begin{quote} \raggedright
$\input{long_equations/L3_n3}$. 
\end{quote}

\subsubsection*{Case VI: $L=3, n=2$}

$\num\bigg( \frac{\partial \ell_{\lambda,E^{(3)}}(\psi)}{\partial b_{1}} \bigg)=$
\begin{quote} \raggedright
$\input{long_equations/L3_n2}$. 
\end{quote}

\section{Basic Concepts from Computational Algebra}
\label{supp:basic_concepts}

\begin{figure}[!hbt]
\centering
\begin{tikzpicture}[scale=1]
    \tikzset{Process0/.style={rectangle,  draw,  text centered, text width=6cm, minimum height=1cm}};
    \tikzset{Process/.style={rectangle,  draw,  text centered, text width=5cm, minimum height=1cm}};
    \tikzset{Process2/.style={rectangle,  draw,  text centered, text width=3.5cm, minimum height=1cm}};
    \tikzset{Process3/.style={rectangle,  draw,  text centered, text width=4cm, minimum height=1cm}};
    \node[Process0](a) at (0,0){System of Algebraic Equations};
    \node[Process](b) at (0,-1.5){Algebraic Variety ($\S$ \ref{sssec:1})};
    \node[Process2](c) at (0,-3){Ideal ($\S$  \ref{sssec:1})};
    \node[Process2](d2) at (-3,-4.5){Saturation ($\S$  \ref{sssec:2})};
    \node[Process2](d3) at (3,-4.5){Elimination ($\S$  \ref{sssec:2})};
    \node[Process3](e) at (0,-6){Gr\"obner Basis ($\S$  \ref{sssec:3})};
    \node[Process](E) at (5,-6){Monomial Ordering ($\S$  \ref{sssec:3})};
    \draw[->] (a) --(b);
    \draw[->] (b) --(c);
    \draw[->] (c) --(d2);
    \draw[->] (c) --(d3);
    \draw[->] (d2) --(e);
    \draw[->] (d3) --(e);
    \draw[->] (E) --(e);
\end{tikzpicture}
\caption{Algebraic concepts.}
\label{tab:concept}
\end{figure}
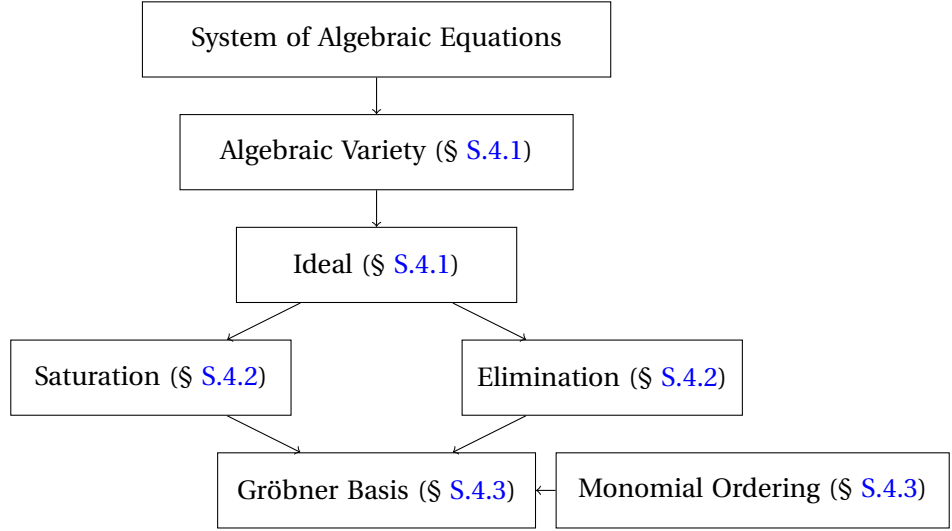

Since algebraic concepts appear frequently in this section, we begin by motivating the necessity of each definition to highlight their relevance. We then introduce foundational notions from computational algebra. 
First, in Section~\ref{sssec:1}, we define \emph{algebraic varieties}, as the solution sets of systems of algebraic equations are precisely such varieties. In the same section, we introduce the concept of \emph{ideals} -- sets of multivariate polynomials -- because solution spaces of algebraic systems can also be characterized as varieties defined by certain ideals. Various operations on ideals in multivariate polynomial rings serve as key tools for solving systems of algebraic equations. 
Next, in Section~\ref{sssec:2}, we discuss two such operations: \emph{saturation} and \emph{elimination}, both of which are central to our method and rely heavily on the theory of Gr\"obner bases. Accordingly, in Section~\ref{sssec:3}, we define Gr\"obner bases, beginning with a review of \emph{monomial orderings}, which are essential for their construction. 
Throughout Sections~\ref{sssec:1}--\ref{sssec:3}, we refer to definitions, results, and proofs presented in the widely accessible textbook by \citet{CLO}, which forms the foundation for our algebraic framework. As illustrated in Figure~\ref{tab:concept}, systems of algebraic equations correspond to algebraic varieties, which in turn correspond to ideals. Operations on these ideals—particularly saturation and elimination—can be systematically computed using Gr\"obner bases, which themselves depend on the choice of a monomial ordering.

\subsection{Algebraic Varieties and Ideals}\label{sssec:1}

Let \(\mathbb{R}[\bm{z}]\) denote the set of all polynomials in \(w\) variables \(\bm{z} = (z_1, z_2, \ldots, z_w)\) with coefficients in the real number field \(\mathbb{R}\). Recall that \(\mathbb{R}[\bm{x}]\) is a commutative ring, which is called the {\em polynomial ring} in variables \(\bm{z}\) with coefficients in \(\mathbb{R}\). Let \(f_1,f_2,\ldots,f_r\) be polynomials in the polynomial ring \(\mathbb{R}[\bm{z}]\). In this section, we consider the following system of algebraic equation:
\begin{align}
     \left\{
    \begin{array}{l}
       0 = f_1(\bm{z}),
       \\
       0 = f_2(\bm{z}),
       \\
       ~~~~\vdots
       \\
       0 = f_r(\bm{z}).
    \end{array}
    \right.
    \label{eq:spe}
\end{align}

Now, let us introduce the {\em algebraic variety} \(\mathbb{V}_{U}(f_1,f_2,\ldots, f_r)\) defined by the polynomials \(f_1,f_2,\ldots,f_r\):
\begin{align*}
   \mathbb{V}_U(f_1,f_2,\ldots,f_r) = \{ \bm{z} \in U^w \mid \eqref{eq:spe} \mbox{ is satisifed} \}.
\end{align*}
Here, while this paper focuses on real solutions and does not address complex solutions, we set \( U = \mathbb{R} \) or \( \mathbb{C} \) to account for certain properties that hold only in the complex space \(\mathbb{C}\). We have to note that all the solutions to Eq. \eqref{eq:spe} form \(\mathbb{V}_U(f_1,f_2,\ldots,f_r)\). In Sections \ref{sssec:1} -- \ref{sssec:3}, we discuss an approach for analyzing the solution space of \eqref{eq:spe} via the algebraic variety \(\mathbb{V}_U(f_1, f_2, \ldots, f_r)\).

Now, let us define the following set of polynomials:
\[
   \ideal{f_1,\, f_2, \, \ldots, \, f_r}
   =
   \left\{
   \sum_{i=1}^r h_i f_i \mid h_i \in \mathbb{R}[\bm{z}]
   \right\},
\]
where \(h_i f_i\) is the product of \(h_i\) and \(f_i\). The set \(\langle f_1,f_2,\ldots,f_r\rangle\) is an ideal in the polynomial ring \(\mathbb{R}[\bm{z}]\). So this set is called the {\em ideal} generated by \(f_1,f_2,\ldots,f_r\). Let \(\mathcal{I} = \ideal{f_1,\, f_2, \, \ldots, \, f_r}\). Now, we introduce the algebraic variety \(\mathbb{V}_U(\mathcal{I})\) defined by $\mathcal{I}$ as like the following:
\begin{align*}
   \mathbb{V}_U(\mathcal{I}) = \{ \bm{z} \in U^w \mid f(\bm{z}) = 0 \; \forall f \in \mathcal{I} \}.
\end{align*}
We have \(\mathbb{V}_U(f_1,f_2,\ldots,f_r) = \mathbb{V}_U(\mathcal{I})\). Therefore, the solution space of \eqref{eq:spe} coincides also with the algebraic variety \(\mathbb{V}_U(\mathcal{I})\).  Various operations can be performed on ideals in multivariate polynomial rings, and these are highly useful for analyzing algebraic varieties. Therefore, in the next section, we introduce two such useful operations: ``saturation'' and ``elimination''.

\subsection{Saturation and Elimination}\label{sssec:2}

In the previous section, we introduced the concept of ideals. First, we present a fundamental property of ideals in the multivariate polynomial ring \(\mathbb{R}[\bm{z}]\), known as Hilbert Basis Theorem \citep[section 2.5, theorem 4]{CLO}.

\begin{theorem}[Hilbert Basis Theorem]\label{thm:hbt}
   Every ideal \(\mathcal{I} \subset \mathbb{R}[\bm{z}]\) has a finite generator set. In other words, \(\mathcal{I} = \ideal{f_1, \, f_2, \, \ldots, \, f_r}\) for some \(f_1, \, f_2, \, \ldots, \, f_r \in \mathbb{R}[\bm{z}]\).
\end{theorem}

By Theorem \ref{thm:hbt}, we can represent any set which is an ideal in the polynomial ring \(\mathbb{R}[\bm{z}]\), by using a finite generator set. Here, we introduce several types of sets that are ideals. The first one is {\em saturations} \citep[section 4.4, definition 8]{CLO}.

\begin{definition}\label{definition:sat}
  If \(\mathcal{I}\) and \(\mathcal{K} \) in the polynomial ring \(\mathbb{R}[\bm{z}]\) are ideals, then the set \(\mathcal{I}:\mathcal{K}^{\infty}\) is 
  \[
      \mathcal{I} : \mathcal{K}^{\infty} = \{ f \in \mathbb{R}[\bm{z}] \mid \forall h \in \mathcal{K} \exists N \geq 0 \mbox{ such that } f h^N \in \mathcal{I}\}
  \]
  and is called the saturation of \(\mathcal{I}\) with respect to \(\mathcal{K}\). Here \(h^N\) denotes the \(N\)-th power of \(h\),
  and \(f h^N\) denotes the product of \(f\) and \(h^N\).
\end{definition}

Saturations are ideals \citep[section 4.4, proposition 9]{CLO}. Therefore, they have finite generator sets by Theorem \ref{thm:hbt}. The computation of a generator set for saturations requires Gr\"obner bases. Accordingly, we define Gr\"obner bases in Section \ref{sssec:3}. We present a fundamental property of algebraic varieties defined by saturations \citep[section 4.4, theorem 10]{CLO}.

\begin{proposition}\label{pro:sat}
  Let \(\mathcal{I}\) and \(\mathcal{K} \) be ideals in the polynomial ring \(\mathbb{R}[\bm{z}]\). Then we have
  \begin{align*}
  \mathbb{V}_{\mathbb{R}}(\mathcal{I} : \mathcal{K}^{\infty}) \supset \overline{\mathbb{V}_{\mathbb{R}}(\mathcal{I}) \setminus \mathbb{V}_{U}(\mathcal{K})},
  \\
  \mathbb{V}_{\mathbb{C}}(\mathcal{I} : \mathcal{K}^{\infty}) = \overline{\mathbb{V}_{\mathbb{C}}(\mathcal{I}) \setminus \mathbb{V}_{\mathbb{C}}(\mathcal{K})}.
  \end{align*}
  Here \(\mathbb{V}_{U}(\mathcal{I}) \setminus \mathbb{V}_{U}(\mathcal{K})\) is the set of elements in \( \mathbb{V}_{U}(\mathcal{I}) \) that are not in \( \mathbb{V}_{U}(\mathcal{K}) \), with $U=\mathbb{R}$ or $U=\mathbb{C}$. 
  Then, \(\overline{\mathbb{V}_{U}(\mathcal{I}) \setminus \mathbb{V}_{U}(\mathcal{K})}\) is the Zariski closure of \(\mathbb{V}_{U}(\mathcal{I}) \setminus \mathbb{V}_{U}(\mathcal{K})\).
\end{proposition}

Roughly speaking, the Zariski closure are the smallest algebraic variety containing a given set. In other words, it is the smallest solution space of a system of algebraic equations that contains a given set. Therefore, if \(\mathcal{K} = \ideal{h_1, \, h_2, \, \ldots, h_s}\) , the algebraic variety \(\mathbb{V}_{\mathbb{C}}(\mathcal{I} : \mathcal{K}^\infty)\) corresponds to the Zariski closure of a set obtained by removing the solution set of the following system from that of System \eqref{eq:spe}:

\begin{align*}
     \left\{
    \begin{array}{l}
       0 = h_1(\bm{z}),
       \\
       0 = h_2(\bm{z}),
       \\
       ~~~~\vdots
       \\
       0 = h_s(\bm{z}).
    \end{array}
    \right.
    \label{eq:spe2}
\end{align*}

That is, we can treat systems of algebraic equations and non-equations. At Section \ref{ssec:3}, we control non-equations by using saturations. Next, we define {\em eliminations} \citep[section 3.1, definition 1]{CLO}.

\begin{definition}\label{definition:elm}
  Let \(\mathcal{I}\) be an ideal in the polynomial ring \(\mathbb{R}[\bm{z}]\). The \(i\)-th elimination of \(\mathcal{I}\), denoted by \(\elim{\mathcal{I}}_i\), is the intersection
  \[
    \elim{\mathcal{I}}_i = \mathcal{I} \cap \mathbb{R}[z_{i+1}, \ldots, z_w].
  \]
\end{definition}

Eliminations are ideal in the polynomial ring \(\mathbb{R}[\bm{z}]\) \citep[section 3.1, exercise 1]{CLO}. Therefore, similarly to saturations, eliminations have finite generator sets. In addition, similarly to saturations, the computations of generator sets of eliminations require Gr\"obner bases. We give a fundamental property of the algebraic varieties for eliminations, which is known as the closure theorem \citep[section 4.4, theorem 4]{CLO}. 

\begin{theorem}[The Closure Theorem]
  Let \(V= \mathbb{V}_{\mathbb{C}}(f_1, \, f_2, \, \ldots, f_r)\) and \(\mathcal{I} = \ideal{f_1, \, f_2, \, \ldots, f_r}\),  and \(\pi_{i} : \mathbb{C}^n \to \mathbb{C}^{n-i} ; (z_1, \, z_2, \, \ldots, \, z_n) \mapsto (z_{i+1}, \, z_{i+2}, \, \ldots, \, z_n)\) be projection onto the last \(n-i\) coordinates. Then 
  \[
    \mathbb{V}_{\mathbb{C}}(\elim{\mathcal{I}}_i) = \overline{\pi_i(V)}.
  \]
\end{theorem}

Recall that our system of algebraic equations contains a Lagrange multiplier at Supplement~\ref{ex2}. We can delete such a Lagrange multiplier by using eliminations. We conclude this section with showing generator sets of the saturations and the eliminations for Supplement~\ref{ex2}. In the following section, we present methods for computing these generator sets.

\subsection{Gr\"obner Basis}\label{sssec:3}

The solution space can be successfully obtained by constructing saturations and eliminations. To obtain simplified systems of algebraic equations, in addition, we need to compute generator sets of saturations and eliminations. Generator sets of saturations and eliminations are obtained by using the theory of Gr\"obner bases. The definition of Gr\"obner bases requires a monomial ordering. Therefore, we define {\em monomial orderings} \citep[section 2.2, definition 7]{CLO}, before giving the definition of Gr\"obner bases. 

Let us denote $\bm{z}^{\bm{a}} = z_1^{a_1} \cdots z_w^{a_w}$ for $\bm{a}=(a_1,\ldots,a_w) \in \mathbb{Z}_{\geq 0}^w$, which is called a monomial, where $\mathbb{Z}_{\geq 0}^w = \{ (\alpha_1,\ldots,\alpha_w) \in \mathbb{Z}^w : \alpha_1, \ldots, \alpha_w \geq 0\}$. Let $M(\bm{z}) = \{ \bm{z}^{\bm{a}} : \bm{a} \in \mathbb{Z}_{\geq 0}^w \}$.
\begin{definition}
  A {\em monomial ordering} $\succ$ on $\mathbb{R}[\bm{z}]$ is a relation on the set $M(\bm{z})$ satisfying that  
  \begin{enumerate}
  \item $\succ$ is a total ordering on $M(\bm{z})$, {that is, $\succ$ satisfies that
    \begin{enumerate}
    \item
      $s \succeq s$ for any $s \in M(\bm{z})$,
    \item
      $s \succeq t$ and $t \succeq s$ implies $s = t$ for any $s,t \in M(\bm{z})$,
    \item
      $s \succeq t$ and $t \succeq u$ implies $s \succeq u$ for any $s,t,u \in M(\bm{z})$, and
    \item
      $s \succ t$ or $s = t$ or $s \prec t$ for any $s,t \in M(\bm{z})$,
    \end{enumerate}}
  \item $\bm{z}^{\bm{a}} \succ \bm{z}^{\bm{b}} \Longrightarrow \bm{z}^{\bm{a}} \bm{z}^{\bm{c}} \succ \bm{z}^{\bm{b}} \bm{z}^{\bm{c}}$ for any $\bm{c} \in \mathbb{Z}_{\geq 0}^m$, and 
  \item any nonempty subset of $M(\bm{z})$ has a smallest element under $\succ$.
  \end{enumerate}
\end{definition}

Next we define {\em leading monomials} \citep[section 2.2, definition 7]{CLO}, which play an important role in the definition of Gr\"obner bases.

\begin{definition}
  Fixing a monomial ordering on $\mathbb{R}[\bm{z}]$ and given a polynomial $f = \sum_{\bm{a} \in \mathbb{Z}_{\geq 0}^w} c_{\bm{a}} \bm{z}^{\bm{a}}$ for $c_{\bm{a}} \in \mathbb{R}$, we call the monomial $\mathrm{LM}(f) = \max(\bm{z}^{\bm{a}} : c_{\bm{a}} \neq 0)$ the leading monomial of $f$.
\end{definition}

We give two examples of monomial orderings. The first one is called a {\em lexicographic order} (or a {\em lex order} for short) \citep[section 2.2, definition 3]{CLO}.  
\begin{definition}[Lexicographic Order]
  Let $\bm{a}, \bm{b} \in \mathbb{Z}_{\geq 0}^m$. We say $\bm{z}^{\bm{a}} \succ_{\mathrm{lex}} \bm{z}^{\bm{b}}$ if the leftmost nonzero entry of the vector difference $\bm{a} - \bm{b}$ is positive.
\end{definition}

Although lex orders often provide Gröbner bases that are easier to solve, it typically entails a higher computational cost. Hence, it is desirable to define a monomial order that reduces such a computational cost. Therefore, as the second one, we define {\em graded reverse lex orders} (or a {\em grevlex orders} for short) \citep[section 2.2, definition 8]{CLO}.
\begin{definition}[Graded Reverse Lex Order]
  Let $\bm{a} = (a_1,\ldots, a_m), \bm{b} = (b_1,\ldots, b_m) \in \mathbb{Z}_{\geq 0}^m$. We say $\bm{z}^{\bm{a}} \succ_{\mathrm{grevlex}} \bm{z}^{\bm{b}}$ if $|\bm{a}| = \sum_{i=1}^{m} a_i > |\bm{b}| = \sum_{i=1}^{m} b_i$ or the rightmost nonzero entry of the vector difference $\bm{a} - \bm{b}$ is negative. 
\end{definition}

Now we define {\em Gr\"obner bases} \citep[section 2.5, definition 5]{CLO}.
\begin{definition}
  Fix a monomial ordering on $\mathbb{R}[\bm{z}]$. A finite subset $G = \{ g_1, \ldots, g_t \}$ of an ideal $\mathcal{I} \subseteq \mathbb{R}[\bm{z}]$ is said to be a {\em Gr\"obner basis} of $\mathcal{I}$ if
  \begin{align*}
    \ideal{\mathrm{LM}(g_1), \ldots, \mathrm{LM}(g_t)}
    =
    \ideal{\mathrm{LM}(f) : f \in \mathcal{I}}
  \end{align*}
\end{definition}

Note that every polynomial ideal has a Gr\"obner basis \citep[section 2.5, corollary 6]{CLO}, and that any Gr\"obner basis of $\mathcal{I}$ generates the ideal $\mathcal{I}$.

For the case $\mathcal{K} = \langle h \rangle \subseteq \mathbb{R}[\bm{z}]$, a generator set of the saturation $\mathcal{I}:\mathcal{K}^{\infty}$ is obtained by using a Gr\"obner basis as follows \citep[section 4.4, theorem 14 (ii)]{CLO}.
\begin{proposition}\label{pro:satcom}
   Let \(\mathcal{I}\) be an ideal in the polynomial ring \(\mathbb{R}[\bm{z}]\), and \(\mathcal{K} = \ideal{h}\) for \(h \in \mathbb{R}[\bm{z}]\). Then we can compute a Gr\"obner basis of the saturation $\mathcal{I}:\mathcal{K}^{\infty}$.
   \begin{enumerate}
   \item let $\tilde{\mathcal{I}} = \langle f_1, \ldots, f_r, 1 - Y h \rangle \subseteq \mathbb{R}[Y, \bm{z}]$, where $Y$ is a new variable,
   \item compute a Gr\"obner basis $\tilde{G}$ of $\tilde{\mathcal{I}}$ with respect to the lex order,
   \item then $G = \tilde{G} \cap \mathbb{R}[\bm{z}]$ is a Gr\"obner basis of the saturation $\mathcal{I}:\mathcal{K}^{\infty}$.
   \end{enumerate}
   In particular, it should be noted that a Gr\"obner basis can be obtained as a generating set.
\end{proposition}

A generator set of the elimination \(\elim{\mathcal{I}}_i\) of an ideal \(\mathcal{I}\) can be computed as follows \citep[section 3.1, theorem 2]{CLO}.
\begin{proposition}\label{pro:elmcom}
   Let \(\mathcal{I}\) be an ideal in the polynomial ring \(\mathbb{R}[\bm{z}]\). Then we can compute a Gr\"obner basis of the \(i\)-th elimination \(\elim{\mathcal{I}}_i\) as the following:
   \begin{enumerate}
   \item compute a Gr\"obner basis $G$ of $\mathcal{I}$ with respect to the lex order,
   \item then $G \cap \mathbb{R}[z_{i+1}, \, z_{i+1}, \, \ldots, z_w]$ is a Gr\"obner basis of the \(i\)-th elimination $\elim{\mathcal{I}}_i$.
   \end{enumerate}
   In particular, as with saturation, it should be noted that a Gr\"obner basis can be obtained as a generating set.
\end{proposition}

In addition, Gr\"obner bases have a good property that plays an important role in solving a system of algebraic equations. The following proposition shows that Gr\"obner bases transform \eqref{eq:spe} into easily solvable problems, without being subject to chance. The following proposition is called the finiteness theorem \citep[section 5.3, theorem 6]{CLO}.

\begin{theorem}[The Finiteness Theorem]\label{finiteness-thm}
  Fix a monomial ordering on $\mathbb{R}[\bm{z}]$ and let $G$ be a Gr\"obner basis of an ideal $\mathcal{I} \subseteq \mathbb{R}[\bm{z}]$.
  Consider the following four statements:
  \begin{enumerate}
  \item\label{finiteness-thm-1} for each $i = 1, \ldots, w$, there exists $t_i \geq 0$ such that $z_i^{t_i} \in \langle \mathrm{LM}(f) : f \in \mathcal{I} \rangle$,
  \item\label{finiteness-thm-2} for each $i = 1, \ldots, w$, there exists $t_i \geq 0$ and $g \in G$ such that $\mathrm{LM}(g) = z_i^{t_i}$,
  \item\label{finiteness-thm-3} for each $i = 1, \ldots, w$, there exists $t_i \geq 0$ and $g \in G$ such that $\mathrm{LM}(g) = z_i^{t_i}$ and $g \in \mathbb{R}[z_i, \ldots, z_w]$, if we fix the lex order,
  \item\label{finiteness-thm-4} the algebraic variety $\mathbb{V}_{U}(\mathcal{I})$ is a finite set.
  \end{enumerate}
  Then the statements \ref{finiteness-thm-1}-\ref{finiteness-thm-3} are equivalent and they all imply the statement \ref{finiteness-thm-4}. An ideal $\mathcal{I}$ satisfying the statement \ref{finiteness-thm-1}, \ref{finiteness-thm-2}, or \ref{finiteness-thm-3} is called a {\em zero-dimensional ideal}. Otherwise it is called a {\em non-zero dimensional ideal}. Furthermore, if $W = \mathbb{C}$, then the statements \ref{finiteness-thm-1}-\ref{finiteness-thm-4} are all equivalent. 
\end{theorem}

The above proposition shows that if we can obtain a Gröbner basis, we can determine whether the algebraic variety is a finite or infinite set, and if the algebraic variety is a finite set, the Gröbner basis with respect to the lex order gives a triangulated representation (statement \ref{finiteness-thm-3} of Proposition \ref{finiteness-thm}).  This representation produces a much simpler system of algebraic equation than the original one; for example, when a given system is a system of linear equations, a Gröbner basis provides a triangular matrix. In general, the computation of a Gr\"obner basis with respect to a grevlex order is more efficient than that with respect to a lex order. Also, if $\mathcal{I}$ is zero-dimensional, we have an efficient algorithm which converts a Gr\"oebner basis from one monomial orderings to another, which is called the FGLM algorithm \citep{FGLM}. Hence, in our algebraic approach, we first compute Gr\"obner bases with respect to the grevlex order, and then convert them to the lex order. 

\begin{remark}\label{rem:zerotest}
  Let \(\mathcal{I}\) be an ideal in \(\mathbb{R}[\bm{z}]\) and \(G\) a Gr\"obner basis of \(\mathcal{I}\). Then, Theorem \ref{finiteness-thm} tell us that
  \(\mathcal{I}\) is zero-dimensional if and only if there exists $t_i \geq 0$ and $g \in G$ such that $\mathrm{LM}(g) = z_i^{t_i}$. That is, as long as we can compute a Gröbner basis, we can determine whether the ideal is zero-dimensional.
\end{remark}

\end{document}